\pdfoutput=1
\documentclass[onecolumn]{tamplate/hexafuture}
\usepackage{fontawesome5}   %
\usepackage{multirow,rotating,url,tabularx}
\AtBeginEnvironment{table}{\renewcommand{\arraystretch}{1.16}}
\newcolumntype{Y}{>{\raggedright\arraybackslash}X}
\usetikzlibrary{arrows.meta,calc,positioning,fit,backgrounds}   %
\definecolor{rsblue}{HTML}{2A6FD0}\definecolor{rsviolet}{HTML}{6A4FC4}\definecolor{rsgreen}{HTML}{1F8F62}
\definecolor{rsgray}{HTML}{6B6A66}\definecolor{rsred}{HTML}{C8402F}\definecolor{rsink}{HTML}{2E2D2A}
\definecolor{skillsblue}{HTML}{2A78D6}\definecolor{baseorange}{HTML}{EB6834}
\newcommand{\qs}[1]{\textquotesingle#1\textquotesingle}   %
\makeatletter
\DeclareRobustCommand\onedot{\futurelet\@let@token\@onedot}
\def\@onedot{\ifx\@let@token.\else.\null\fi\xspace}
\makeatother

\def\etal{\emph{et al}\onedot}

\title{HexaShell: A Universal Harness for Robot Coding Agents in Simulation and the Real World}

\author{
\begin{center}
    Hongcheng Gao$^{*\dagger}$ \quad Jianshu Zeng$^{*}$ \quad Alex Zhou$^{*}$ \quad Shijia Ge \quad Xuan Shangguan \quad Yazhe Wang  
    \\[5pt]
    \quad Yexing Wan \quad Zelin Zheng \quad Jay Zhu \quad Xiping Li \quad Di Wu \quad Yongli Wang 
    \\[5pt]    
    \quad Lingyu He \quad Zhiqi Jia \quad Sihang Wu \quad Xiao He$^{\dagger}$
    \\[8pt]
    $^{*}$Equal Contribution \qquad $^{\dagger}$Corresponding Author
    \\[8pt]
    {\sffamily\small\color{hexaprimary}\faGlobe\enspace\href{https://hexafuture.ai}{hexafuture.ai}}
    \vspace{-0.5cm}
\end{center}
}

\FloatBarrier
\abstract{Agents built on vision language models (VLMs) now control robots by writing code, but the strongest of them refine task-specific code over many rounds of trial and revision. Such code tends to overfit the benchmark on which it was tuned, can exploit the benchmark's shortcuts, and is costly to carry over to a new benchmark or robot. We present HexaShell, a thin, task-agnostic harness designed to test how far general tools can go instead. We design a simple and complete set of 14 basic tools, 4 for motion, 5 for perception and 5 for state, which the agent calls in one Python cell per turn. Perception pairs SAM~3, which segments an object from a few words or a click, with NumPy and OpenCV geometry that turns its outlines in several cameras, or a depth image, into its position, size and grasp angle. The API, agent loop and sandbox are shared across all tasks; a new simulator or robot needs only an adapter of 350 to 1,450 lines and a prompt describing its world, and the only task-specific input is an optional skill of about a page of guidance. The same code runs on four simulated benchmarks, built on three simulators with two kinds of arm, and on real robots. With GPT-6-Astra and no task-specific input, HexaShell reaches success rates of 69.9\% on RoboDojo, 59.3\% on RoboCasa365, 78.3\% on LIBERO-PRO and 82.2\% on RoboTwin~2.0. With task skills, it reaches 85.0\% (score 88.8), 85.6\%, 96.7\% and 96.8\%, respectively, above the best published result we are aware of on each benchmark. The RoboCasa365 result is listed first on its leaderboard. On real dual-arm robots, HexaShell succeeds in 76 of 90 episodes across nine tasks.}

\begin{document}
\maketitle
\justifying
\setlength{\parindent}{0pt}

\vspace{-8pt}
{\setlength{\intextsep}{0pt}\begin{figure}[H]
\centering
\captionsetup{skip=5pt}
\hbox to \linewidth{\hfil$\vcenter{\hbox{\includegraphics{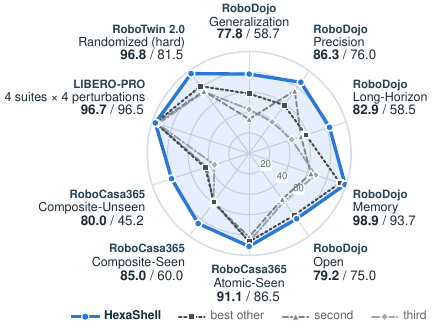}}}$\hskip 5mm$\vcenter{\hbox{%
\begin{tikzpicture}[x=1mm, y=1mm, font=\sffamily\fontsize{6.4}{7.6}\selectfont, text=rsink,
  card/.style={rounded corners=2.5pt, line width=0.55pt, draw=#1, fill=#1!5},
  ttl/.style={font=\sffamily\bfseries\fontsize{8}{9.6}\selectfont, text=#1!80!black, anchor=base west, inner sep=0pt},
  sub/.style={anchor=base west, inner sep=0pt, text=rsink!85},
  img/.style={anchor=north west, inner sep=0pt},
  tag/.style={font=\sffamily\fontsize{5.8}{6.8}\selectfont, anchor=base, inner sep=0pt, text=rsink!85},
  val/.style={font=\sffamily\bfseries\fontsize{6.8}{7.2}\selectfont, anchor=base, inner sep=0pt, text=#1!80!black},
  big/.style={font=\sffamily\bfseries\fontsize{9}{10}\selectfont, anchor=base, inner sep=0pt, text=#1!80!black}]
\draw[card=rsblue] (0,0) rectangle (73,-15.4);
\node[ttl=rsblue] at (2,-3.95) {Simple and complete general tools};
\foreach \n/\c [count=\i from 0] in {4/motion tools, 5/perception tools, 5/state tools} {
  \draw[rsblue!55, fill=white, line width=0.35pt, rounded corners=1pt] (2+18.2*\i,-5.6) rectangle (19+18.2*\i,-13.8);
  \node[anchor=base, inner sep=0pt, font=\sffamily\bfseries\fontsize{11}{11}\selectfont, text=rsblue!80!black] at (10.5+18.2*\i,-9.6) {\n};
  \node[anchor=base, inner sep=0pt, font=\sffamily\fontsize{6.2}{7}\selectfont] at (10.5+18.2*\i,-12.4) {\c};
}
\node[anchor=base, inner sep=0pt, font=\sffamily\bfseries\fontsize{13}{13}\selectfont, text=rsblue!80!black] at (63.4,-9.8) {14};
\node[anchor=base, inner sep=0pt, font=\sffamily\bfseries\fontsize{7}{8}\selectfont, text=rsblue!80!black] at (63.4,-12.6) {basic tools};
\begin{scope}[yshift=-16.6mm]
\draw[card=rsviolet] (0,0) rectangle (73,-19.0);
\node[ttl=rsviolet] at (2,-3.95) {Cross-embodiment harness};
\node[sub, anchor=base, font=\sffamily\fontsize{6.2}{7.4}\selectfont] at (36.5,-7.0) {14 basic tools \textbf{\textcolor{rsviolet!80!black}{+ 1 general prompt}} per benchmark (with a thin adapter):};
\foreach \f/\t/\v [count=\i from 0] in {set_dojo/RoboDojo/69.9\%, set_casa/RoboCasa365/59.3\%, set_libero/LIBERO-PRO/78.3\%,
                                         set_twin/RoboTwin~2.0/82.2\%, set_real/Real robots/76 of 90} {
  \begin{scope}   %
    \clip (3.2+13.8*\i,-8.1) rectangle (14.6+13.8*\i,-13.8);
    \node[inner sep=0pt] at (8.9+13.8*\i,-10.95) {\includegraphics[height=5.7mm]{fig/img/\f.jpg}};
  \end{scope}
  \node[tag] at (8.9+13.8*\i,-15.7) {\t};
  \node[val=rsviolet] at (8.9+13.8*\i,-17.9) {\v};
}
\end{scope}
\begin{scope}[yshift=-36.8mm]
\draw[card=rsgreen] (0,0) rectangle (73,-17.4);
\node[ttl=rsgreen] at (2,-3.95) {Lightweight task skills};
\draw[rsgreen!55, fill=white, line width=0.35pt, rounded corners=0.8pt] (2,-5.6) rectangle (15.4,-16.0);
\node[anchor=north west, inner sep=0pt, align=left, font=\fontfamily{lmtt}\fontsize{3.4}{3.8}\selectfont, text=rsink!80] at (2.7,-6.3)
  {\textcolor{rsgreen!70!black}{\#\# Judging}\\- 100 only if ...\\\textcolor{rsgreen!70!black}{\#\# Flow}\\1. measure ...\\\textcolor{rsgreen!70!black}{\#\# Failures}\\- tilt 3 deg: 0/13};
\node[anchor=north west, inner sep=0pt, align=left, text width=53.6mm, text=rsink!85, font=\sffamily\fontsize{6.2}{7.4}\selectfont] at (17.4,-5.3)
  {14 tools + 1 general prompt \textbf{\textcolor{rsgreen!80!black}{+ 1 task skill}} per task (a page of guidance), drawn from runs on a few seeds:};
\foreach \v/\b [count=\i from 0] in {85.0\%/RoboDojo, 85.6\%/RoboCasa365, 96.7\%/LIBERO-PRO, 96.8\%/RoboTwin~2.0} {
  \node[big=rsgreen] at (24.1+13.4*\i,-13.5) {\v};
  \node[tag] at (24.1+13.4*\i,-16.1) {\b};
}
\end{scope}
\end{tikzpicture}}}$\hfil}
\caption{\textbf{HexaShell on four simulated benchmarks and real robots.} Left: success rates (\%) of HexaShell with task skills (bold) and, after the slash, of the best of the three strongest other methods on each benchmark (\cref{app:comparisons}). Right: its three components.}
\label{fig:teaser}
\end{figure}
}
\clearpage   %
\FloatBarrier
\section{Introduction}
\label{sec:intro}

Vision language models (VLMs) that write code can control robots: given a task and camera images, the model writes a program against the robot's interface, executes it and observes the outcome \cite{liang2023code,singh2022progprompt,huang2023voxposer}. Used in this way, frontier models now rival learned policies on manipulation benchmarks without a single demonstration \cite{zhang2026unexpected,si2026pyrua,chen2026showharness}. Their performance, however, depends to a large extent on the harness that connects the model to the robot. With the same model, GPT-6-Astra, the RoboDojo team's agent reaches a success rate of 22.5\% on RoboDojo \cite{zhang2026unexpected}, whereas ours reaches 69.9\%.

Building such a harness poses two challenges. The first is spatial grounding. The model perceives objects as pixels, yet every motion requires a position accurate to about a centimetre, and VLMs judge such distances poorly \cite{chen2024spatialvlm,cheng2024spatialrgpt}. The second is task knowledge: what counts as completion, how deep a plug must be seated, which arm should move first. Existing approaches place this burden at different levels. Vision-language-action models learn both from demonstrations collected for each task and robot \cite{brohan2023rt2,kim2024openvla,black2024pi0}. Code-as-policies systems expose a fixed set of hand-written primitives, which the model composes into a program for each instruction \cite{liang2023code,huang2023voxposer}. The strongest agent systems instead accumulate tools, programs or memories for the tasks they encounter, often through many rounds of trial and revision \cite{mmlab2026physicalrsi,li2026embodiedevo,sygkounas2026memento,lu2026aspire}. Such task-specific code is costly to develop and difficult to audit, and it tends to fit the scenes on which it was developed. Our earlier exploration, HexaShell-TaskTools (HexaShell-Preview), illustrates the problem: an optimizer agent wrote about 78,000 lines of task tools for RoboDojo over 769 rounds, yet once frozen, these tools passed only 36\% of the development episodes on which they had been built (\cref{sec:knowledge}).

In this work we study how far a harness can go when nothing elaborate is written for any individual task. Our approach rests on an observation about where coding agents fail: most of what they lack is not specific to a task. They need to know where objects are, to verify a motion before committing to it, and to receive clear feedback when something goes wrong, and these needs are common to all tasks and robots. The task-specific remainder is small: a success condition that the instruction leaves implicit, a parameter that only contact reveals, an order of steps. This suggests a division of labour in which general capabilities are implemented once, as shared and tested code, while task knowledge is conveyed in a short natural-language skill.

We present HexaShell, a harness built on this division (\cref{fig:method}). A single robot API, agent loop and sandbox serve every task in every setting; at each turn the model writes one Python cell against the API and reads its output. Perception is decomposed into segmentation and localization. SAM~3 \cite{carion2025sam3} determines which pixels belong to an object, and geometry implemented with NumPy and OpenCV determines where those pixels lie in the world, by matching outlines across calibrated cameras or back-projecting a depth image, so that the VLM never has to estimate a distance. The API is designed with a language model as its user: motions can be checked before they are executed, results report problems first, and errors state what to change. Porting to a new simulator or robot requires only an adapter below the API and the text of the prompt. Task-specific input is limited to an optional skill, about a page of guidance that an analysis agent drafts from the records of earlier runs and a person reviews.

We evaluate HexaShell on four simulated benchmarks, RoboDojo \cite{robodojo2025}, RoboCasa365 \cite{nasiriany2026robocasa365}, LIBERO-PRO \cite{zhou2025liberopro} and RoboTwin~2.0 \cite{chen2025robotwin}, which together span three simulators, two kinds of arm, fixed and wheeled bases, and cameras with and without depth, as well as on real dual-arm PiPER-X robots. Without any task-specific input, HexaShell reaches a success rate of 69.9\% on RoboDojo, above every entry of its leaderboard, and in nine episodes out of ten it writes no vision code of its own. With task skills, it reaches 85.0\% on RoboDojo, 85.6\% on RoboCasa365, 96.7\% on LIBERO-PRO and 96.8\% on RoboTwin~2.0, above the best published result we are aware of on each benchmark (\cref{fig:teaser}). The gains from skills stem from a few facts per task. The remaining failures are dominated by grasps that slip, which originate below the API and lie beyond the reach of any skill.

Our main contributions are as follows.
\begin{itemize}
    \item \textbf{Simple and complete general tools.} We design a simple and complete set of general tools, shared by all tasks, through which the model acts by writing Python: \texttt{goto} moves an arm, \texttt{find} segments objects, \texttt{measure} locates them in 3D and \texttt{state} reports the arms and the remaining steps. Perception separates segmentation by SAM~3 from geometric localization, so that the model receives metric 3D positions with error estimates instead of judging distances itself, and every motion can be checked in a dry run before it consumes steps. With nothing written for any task, GPT-6-Astra in HexaShell reaches a success rate of 69.9\% on RoboDojo, above every entry of its leaderboard, whereas the same model in the RoboDojo team's agent reaches 22.5\% (\cref{sec:main}).
    \item \textbf{A single harness across benchmarks and embodiments.} The same API, loop and sandbox run unchanged on the four simulated benchmarks and on real robots, where HexaShell succeeds in 76 of 90 episodes across nine tasks; each setting requires only an adapter of 350 to 1,450 lines and its own prompt text. Without task skills, HexaShell reaches 69.9\% on RoboDojo, 59.3\% on RoboCasa365, 78.3\% on LIBERO-PRO and 82.2\% on RoboTwin~2.0 (\cref{sec:main,sec:universal}).
    \item \textbf{Task knowledge as short natural-language skills.} We show that what general tools cannot supply, namely the success condition, parameters that only contact reveals and the order of steps, fits in about a page of guidance per task, which an analysis agent drafts from the records of earlier runs. With these skills, HexaShell reaches 85.0\% on RoboDojo, 85.6\% on RoboCasa365, 96.7\% on LIBERO-PRO and 96.8\% on RoboTwin~2.0, above the best published result we are aware of on each benchmark; on RoboDojo, of the 1,020 episodes outside the Open dimension, 234 succeed only with skills and 37 only without (\cref{fig:teaser,sec:knowledge}).
\end{itemize}

\FloatBarrier
\section{HexaShell}
\label{sec:method}

HexaShell separates what is shared from what is specific. A single robot API, agent loop and sandbox, written and tested once, serve every task in every setting, and their perception and motion tools are general enough that the agent rarely needs more (\cref{sec:loop,sec:perception}). Setting-specific input is limited to an adapter below the API and the text of the prompt (\cref{sec:adapters}); task-specific input is limited to an optional skill, about a page of guidance on the success condition, the steps that work and a few parameters from earlier runs (\cref{sec:skills}). \Cref{sec:experiments} evaluates the general tools alone, the transfer of the harness across settings, and the effect of the skills.

\begin{figure}[t]
\centering
\def\methodkey#1{{\fboxsep=0pt\fboxrule=0.5pt$\vcenter{\hbox{\fcolorbox{#1}{#1!12}{\rule{0pt}{2.6mm}\rule{2.6mm}{0pt}}}}$}}%
\resizebox{\linewidth}{!}{%
\begin{tikzpicture}[
  x=1mm, y=1mm, font=\sffamily\fontsize{8.4}{10.1}\selectfont, text=rsink, >={Stealth[length=3.8pt,width=3.4pt]},
  box/.style={rounded corners=3pt, line width=0.6pt, draw=#1, fill=white},
  ttl/.style={font=\sffamily\bfseries\fontsize{9.4}{11.3}\selectfont, text=#1!80!black, anchor=north, inner sep=0pt},
  chip/.style={rounded corners=1.8pt, draw=#1!60, fill=#1!6, line width=0.45pt, inner xsep=2.8pt, inner ysep=2.2pt, align=left, anchor=north west},
  code/.style={rounded corners=1.4pt, draw=#1!55, fill=white, line width=0.4pt, inner xsep=2.4pt, inner ysep=1.8pt, font=\ttfamily\fontsize{8}{9.6}\selectfont, anchor=base west},
  txt/.style={anchor=north west, inner sep=0pt, align=left},
  sub/.style={font=\sffamily\fontsize{7.7}{9.2}\selectfont},
  small/.style={font=\sffamily\fontsize{8}{9.6}\selectfont},
  arr/.style={->, line width=0.65pt, draw=rsink!70},
  lbl/.style={font=\sffamily\fontsize{7.8}{9.4}\selectfont, text=rsink!80, inner sep=1.2pt},
  side/.style={font=\sffamily\bfseries\fontsize{8.8}{10}\selectfont, rotate=90, anchor=south, inner sep=0pt},
]
\draw[box=rsink!55] (4,0) rectangle (64,-30);
\node[ttl=rsink] at (34,-2.2) {What the model reads};
\begin{scope}[every node/.append style={font=\sffamily\fontsize{7.8}{9.4}\selectfont, text width=53.4mm}]
\node[chip=rsviolet] at (6.2,-8.4) {\textbf{System prompt}\enspace six parts, per benchmark};
\node[chip=rsgray] at (6.2,-15.6) {\textbf{Task card}\enspace instruction, budget, start images};
\node[chip=rsgreen] at (6.2,-22.8) {\textbf{Skill}\enspace about a page per task (optional)};
\end{scope}

\draw[box=rsblue] (69,0) rectangle (100,-30);
\node[ttl=rsblue] at (84.5,-2.2) {VLM};
\node[txt] at (71.2,-7.8) {e.g.\ \textbf{GPT-6-Astra}};
\node[txt] at (71.2,-11.9) {one Python cell\\per turn};
\node[code=rsblue] at (71.2,-22.3) {python(code)};
\node[code=rsblue] at (71.2,-27.2) {finish(status)};

\draw[box=rsblue] (110,0) rectangle (160,-30);
\node[ttl=rsblue] at (135,-2.2) {Sandboxed Python worker};
\node[txt, font=\sffamily\fontsize{8.4}{12.5}\selectfont] at (112.2,-7.8) {persistent session,\\no network or subprocesses,\\workspace files only ...};
\coordinate (c) at (112.2,-26.0);
\foreach \c in {numpy,cv2,math,json} {
  \node[code=rsblue] (cc) at (c) {\c};
  \coordinate (c) at ($(cc.base east)+(1.5,0)$);
}
\node[anchor=base west, inner sep=0pt, small] at (c) {\dots};

\draw[arr] (64,-15) -- (69,-15);
\draw[arr] (100,-10) -- node[lbl, above] {cell} (110,-10);
\draw[arr] (110,-20) -- node[lbl, below] {output} (100,-20);

\draw[box=rsblue] (4,-36) rectangle (100,-63.2);
\node[ttl=rsblue] at (52,-38.2) {Perception};
\node[anchor=north west, inner sep=0pt] at (6.2,-43.6) {\includegraphics[width=19.5mm]{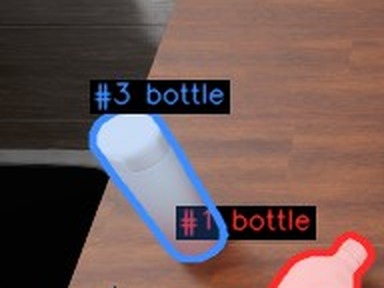}};
\begin{scope}[every node/.append style={font=\sffamily\fontsize{7.6}{9.1}\selectfont}]
\node[chip=rsblue, text width=37.5mm] (sam) at (28,-43.0) {\textbf{SAM 3}\enspace which pixels\\[-0.3pt]{\sffamily\fontsize{7}{8.4}\selectfont from words, clicks or a box}};
\node[chip=rsblue, text width=37.5mm] (geo) at (28,-53.8) {\textbf{Geometry}\enspace where in the world\\[-0.3pt]{\sffamily\fontsize{7}{8.4}\selectfont NumPy, OpenCV; views or depth}};
\end{scope}
\draw[arr] (sam.south) -- (geo.north);
\node[chip=rsblue, text width=25mm, fill=white, font=\ttfamily\fontsize{7}{8.2}\selectfont, anchor=west] (thing) at (71,-55.0) {Thing\#3 \qs{bottle}\\at (-0.28, -0.05, 0.85)\\size 0.06$\times$0.06$\times$0.17\\2 views, $\pm$2 mm};
\draw[arr] (geo.east) -- (geo.east -| thing.west);

\draw[box=rsblue] (110,-36) rectangle (160,-63.2);
\node[ttl=rsblue] at (135,-38.2) {Robot API};
\foreach \y/\what/\calls in {-46.2/move/{goto,gripper,home,wait}, -53.0/see/{look,find,measure}, -59.8/points/{to\_world,to\_pixel}} {
  \node[anchor=base west, inner sep=0pt, font=\sffamily\bfseries\fontsize{8.4}{10}\selectfont, text=rsink] at (112.0,\y) {\what};
  \coordinate (c) at (122.2,\y);
  \foreach \c in \calls {
    \node[code=rsblue, inner xsep=1.8pt] (cc) at (c) {\c};
    \coordinate (c) at ($(cc.base east)+(0.9,0)$);
  }
}
\draw[arr] (128,-30) -- node[lbl, left] {\texttt{robot.*} calls} (128,-36);
\draw[arr] (150,-36) -- node[lbl, right] {results} (150,-30);
\draw[arr] (110,-45.5) -- node[lbl, above] {images} (100,-45.5);
\draw[arr] (100,-56.5) -- node[lbl, below] {Things} (110,-56.5);

\draw[<->, line width=0.65pt, draw=rsviolet] (135,-63.2) -- (135,-66.8);
\draw[rounded corners=2pt, line width=0.6pt, draw=rsviolet, fill=white] (4,-66.8) rectangle (160,-73.0);
\node[text=rsviolet!80!black] at (82,-69.9) {\textbf{Adapters}, one per setting: motions out; images, camera models and arm state back};
\foreach \f/\name/\robot/\what [count=\i from 0] in {
  set_dojo/RoboDojo/{two X5 arms}/{Isaac Sim, RGB},
  set_twin/RoboTwin~2.0/{two X5 arms}/{SAPIEN, RGB},
  set_libero/LIBERO-PRO/{Panda}/{MuJoCo, RGB-D},
  set_casa/RoboCasa365/{mobile Panda}/{MuJoCo, RGB-D},
  set_real/Real robot/{two PiPER-X arms}/{RGB-D}} {
  \node[anchor=north west, inner sep=0pt, minimum width=29.6mm, minimum height=14.8mm] (t\i) at (4+31.6*\i,-75.6) {};
  \begin{scope}
    \clip (t\i.north west) rectangle (t\i.south east);
    \node[inner sep=0pt] at (t\i.center) {\includegraphics[height=14.8mm]{fig/img/\f.jpg}};
  \end{scope}
  \draw[line width=0.65pt, draw=rsviolet] (t\i.north |- 0,-73.0) -- (t\i.north);
  \node[anchor=north, inner sep=0pt, align=center] at ($(t\i.south)+(0,-1.4)$) {{\fontsize{8}{9.6}\selectfont\textbf{\name}}\\[-0.3pt]{\fontsize{7.7}{9.2}\selectfont\robot}\\[-0.6pt]{\fontsize{7.7}{9.2}\selectfont\what}};
}

\node[small, inner sep=0pt] at (82,-105.4) {%
  \methodkey{rsblue}\hspace{1.3mm}same in every setting\hspace{6.5mm}%
  \methodkey{rsviolet}\hspace{1.3mm}written once per setting\hspace{6.5mm}%
  \methodkey{rsgreen}\hspace{1.3mm}written once per task\hspace{6.5mm}%
  \methodkey{rsgray}\hspace{1.3mm}official task instruction};
\node[side, text=rsblue!80!black] at (1.2,-15) {agent};
\node[side, text=rsblue!80!black] at (1.2,-49.6) {host};
\node[side, text=rsviolet!80!black] at (1.2,-87.5) {settings};
\begin{scope}[on background layer]
\fill[rsblue!4, rounded corners=4pt] (-2.4,2) rectangle (162,-32.6);
\fill[rsblue!4, rounded corners=4pt] (-2.4,-33.4) rectangle (162,-64.8);
\fill[rsviolet!4, rounded corners=4pt] (-2.4,-65.6) rectangle (162,-108.6);
\end{scope}
\end{tikzpicture}}
\caption{\textbf{Overview.} At each turn the model writes one Python cell, which runs in a sandbox. Its \texttt{robot} calls go to the host, where perception locates objects and an adapter turns motions into the actions of each setting.}
\label{fig:method}
\end{figure}

\Cref{fig:method} gives an overview. The agent loop exchanges messages with the model, whose Python cells run in a sandbox and call the robot API; perception and motion run behind the API, outside the sandbox. The model reads three texts: a system prompt, a task card and, when skills are enabled, the task's skill. Three principles guide the design. First, the model makes every decision, while the API executes one request at a time and reports the outcome. Second, the agent observes the scene only through what a real robot also has: camera images (with depth where the benchmark renders it), camera calibration and the state of the arms. Third, success is judged solely by the benchmark's own checker.

\subsection{Agent loop and robot API}
\label{sec:loop}
\label{sec:api}

The model interacts with the robot through a minimal tool interface. It executes Python cells in a persistent session, each returning the printed output, the value of the last expression, any exception and the requested images, and it ends the episode with a stated reason (\cref{app:loop}). The simulation is paused while the model deliberates, so perception and reasoning are free, and only motion consumes the benchmark's own step budget.

Inside a cell, the robot is exposed as a single object, \texttt{robot}. \Cref{tab:api} lists the calls shared by the four simulated benchmarks, in three groups: motion, perception and state. RoboCasa365 and LIBERO-PRO add a few calls for their robots (\cref{app:api-conv}), the real robots do not yet provide \texttt{find}, \texttt{measure} or \texttt{things}, and \cref{app:api} gives the full reference. Motions are specified as poses of an arm's grasp point rather than as joint commands. Any motion can first be executed as a dry run, which reports reachability, path, step cost and proximity to the other arm or to a measured object without moving the robot, so that the model can test a plan before it spends steps. Motions meant to meet resistance, such as presses and insertions, are flagged as contact motions and are not stopped early.

\begin{table}[!htbp]
\centering
\caption{\textbf{Robot API.} The calls shared by the four simulated benchmarks; only motion calls consume simulation steps.}
\label{tab:api}
\small
\begin{tabularx}{\linewidth}{@{}l>{\raggedright\arraybackslash}X@{}}
\toprule
Call & Effect \\
\midrule
\multicolumn{2}{@{}l}{\emph{Motion (consumes steps)}} \\
\texttt{goto(arm, xyz, pitch, yaw, roll, \dots)} & move the grasp point to a pose; \texttt{dry\_run} only checks the motion \\
\texttt{gripper(arm, opening)} & open, close or set an opening between 0 and 1 \\
\texttt{home(arm)}, \texttt{wait(n)} & return to the start pose; hold still for $n$ steps \\
\midrule
\multicolumn{2}{@{}l}{\emph{Perception (no steps)}} \\
\texttt{look(cams, \dots)} & attach the current images, with optional overlays \\
\texttt{find(cam, words, \dots)} & segment objects with SAM~3, also from points or a box \\
\texttt{measure(target)} & 3D centre, size and grasp yaw of an object, with an error estimate \\
\texttt{to\_world(cam, (x, y))} & world point seen at a pixel, with its error \\
\texttt{to\_pixel(xyz, cam)} & pixel at which a world point appears \\
\midrule
\multicolumn{2}{@{}l}{\emph{State (no steps)}} \\
\texttt{task}, \texttt{state()}, \texttt{things} & instruction and steps left; arms and grippers; objects measured so far \\
\texttt{image(cam)}, \texttt{camera(cam)} & raw image; camera intrinsics and pose \\
\bottomrule
\end{tabularx}
\end{table}

The API is designed with a language model as its reader. Each result prints as a short summary with warnings first (\cref{lst:results}) and remains accessible as a structured object in code. Errors are raised as typed exceptions whose messages prescribe a remedy. An unreachable pose leaves the robot untouched and comes with a reachable alternative; a stalled motion or a lost grasp aborts the rest of the cell, so that the model observes the scene again before it continues.

\begin{figure}[t]
\centering
\begin{minipage}{0.96\linewidth}
\begin{lstlisting}[basicstyle=\ttfamily\scriptsize,frame=single,columns=fullflexible,keepspaces=true]
Move left: at (0.112, -0.083, 0.861) pitch 90 yaw 12 roll 0  +-1 mm 0deg  14 steps (line)  [986 steps left]
Grip right: closed on something ~3.4 cm wide (holding)  6 steps  [980 steps left]
Thing#3 'red cup'  center (0.112, -0.083, 0.801)  top 0.836  size 0.071x0.069x0.071 round  grip_yaw any
        on table  2 views (head, left_wrist) +-3mm
Mask#7 head 'red cup' score 0.91  centroid (412, 233)  box (391, 210)-(433, 258)  1520 px
Spot (0.112, -0.083, 0.836) from head (412, 220) on the same surface in head and left_wrist  (~1.6 mm per pixel)
\end{lstlisting}
\end{minipage}
\caption{\textbf{Printed results}, as shown in the RoboDojo prompt: a motion, a grasp, a measured object, a mask and a world point. Warnings are printed first.}
\label{lst:results}
\end{figure}

\subsection{Grounded perception}
\label{sec:perception}

Perception addresses two questions: which pixels show an object, and where those pixels lie in the world. HexaShell answers the first with a learned segmentation model and the second with geometry implemented in NumPy and OpenCV \cite{bradski2000opencv}, so that the VLM never has to judge a distance itself. \texttt{find} queries a SAM~3 service, one per machine \cite{carion2025sam3}, with a few words, pixels marked on or off the object, or a box, and returns outlines (masks) with their scores.

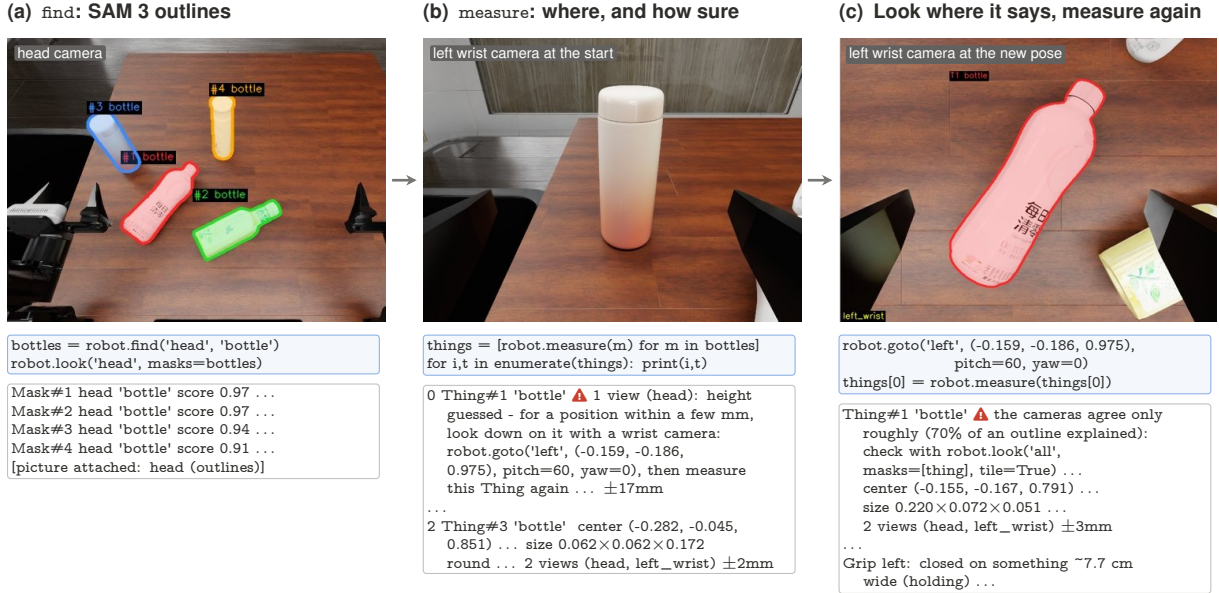
\begin{figure}[t]
\centering
\begin{tikzpicture}[
  x=1mm, y=1mm, font=\sffamily\fontsize{6.5}{7.8}\selectfont, text=rsink, >={Stealth[length=4pt,width=4pt]},
  head/.style={font=\sffamily\bfseries\fontsize{7.5}{9}\selectfont, anchor=base west, inner sep=0pt},
  img/.style={anchor=north west, inner sep=0pt},
  code/.style={rounded corners=1.2pt, draw=rsblue!55, fill=rsblue!5, line width=0.4pt, inner xsep=1.4pt, inner ysep=1.3pt,
               font=\ttfamily\fontsize{6}{7}\selectfont, anchor=north west, align=left, text width=48.2mm},
  outp/.style={rounded corners=1.2pt, draw=rsink!30, fill=white, line width=0.4pt, inner xsep=1.4pt, inner ysep=1.3pt,
              font=\ttfamily\fontsize{6}{7}\selectfont, anchor=north west, align=left, text width=48.2mm},
  arr/.style={->, line width=1pt, draw=rsink!65},
]
\newcommand{\ii}{\hspace*{1.2em}}
\newcommand{\warn}{{\normalfont\fontsize{5}{5}\selectfont\color{rsred}\faExclamationTriangle}}
\node[head] at (0,0) {(a)\enspace\textmd{\texttt{find}}: SAM~3 outlines};
\node[img] (ia) at (0,-2.6) {\includegraphics[width=50mm]{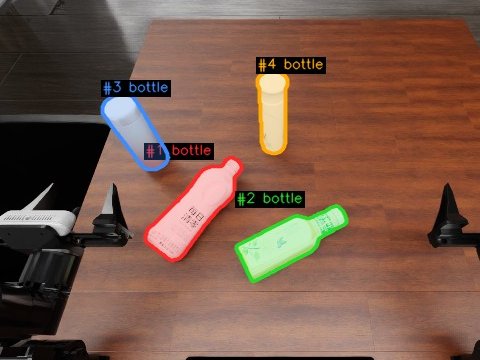}};
\node[code] (ca) at ($(ia.south west)+(0,-1.6)$) {bottles = robot.find(\qs{head}, \qs{bottle})\\robot.look(\qs{head}, masks=bottles)};
\node[outp] at ($(ca.south west)+(0,-1.2)$) {Mask\#1 head \qs{bottle} score 0.97 \dots\\Mask\#2 head \qs{bottle} score 0.97 \dots\\Mask\#3 head \qs{bottle} score 0.94 \dots\\Mask\#4 head \qs{bottle} score 0.91 \dots\\{[}picture attached: head (outlines){]}};
\node[head] at (55,0) {(b)\enspace\textmd{\texttt{measure}}: where, and how sure};
\node[img] (ib) at (55,-2.6) {\includegraphics[width=50mm]{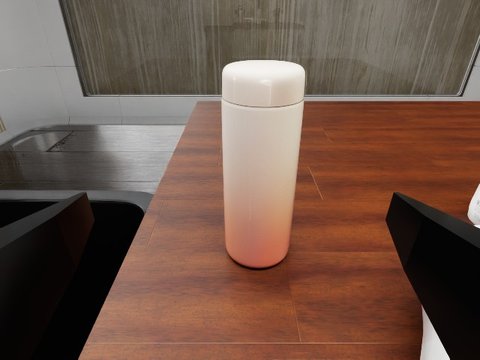}};
\node[code] (cb) at ($(ib.south west)+(0,-1.6)$) {things = [robot.measure(m) for m in bottles]\\for i,t in enumerate(things): print(i,t)};
\node[outp] at ($(cb.south west)+(0,-1.2)$) {0 Thing\#1 \qs{bottle} \warn\ 1 view (head): height\\\ii guessed - for a position within a few mm,\\\ii look down on it with a wrist camera:\\\ii robot.goto(\qs{left}, (-0.159, -0.186,\\\ii 0.975), pitch=60, yaw=0), then measure\\\ii this Thing again \dots\ $\pm$17mm\\\dots\\2 Thing\#3 \qs{bottle} \ center (-0.282, -0.045,\\\ii 0.851) \dots\ size 0.062$\times$0.062$\times$0.172\\\ii round \dots\ 2 views (head, left\_wrist) $\pm$2mm};
\node[head] at (110,0) {(c)\enspace Look where it says, measure again};
\node[img] (ic) at (110,-2.6) {\includegraphics[width=50mm]{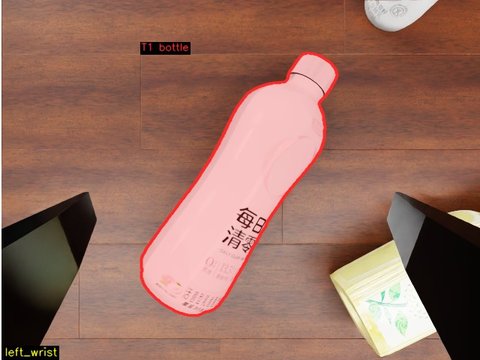}};
\node[code] (cc) at ($(ic.south west)+(0,-1.6)$) {robot.goto(\qs{left}, (-0.159, -0.186, 0.975),\\\hspace*{6.6em}pitch=60, yaw=0)\\things[0] = robot.measure(things[0])};
\node[outp] at ($(cc.south west)+(0,-1.2)$) {Thing\#1 \qs{bottle} \warn\ the cameras agree only\\\ii roughly (70\% of an outline explained):\\\ii check with robot.look(\qs{all},\\\ii masks=[thing], tile=True) \dots\\\ii center (-0.155, -0.167, 0.791) \dots\\\ii size 0.220$\times$0.072$\times$0.051 \dots\\\ii 2 views (head, left\_wrist) $\pm$3mm\\\dots\\Grip left: closed on something \textasciitilde7.7 cm\\\ii wide (holding) \dots};
\foreach \n/\t in {ia/head camera, ib/left wrist camera at the start, ic/left wrist camera at the new pose}
  \node[anchor=north west, fill=rsink!75, text=white, font=\sffamily\fontsize{5.5}{6}\selectfont, inner xsep=1.2pt, inner ysep=0.9pt, rounded corners=0.8pt] at ($(\n.north west)+(0.9,-0.9)$) {\t};
\draw[arr] ($(ia.east)+(0.8,0)$) -- ($(ib.west)+(-0.8,0)$);
\draw[arr] ($(ib.east)+(0.8,0)$) -- ($(ic.west)+(-0.8,0)$);
\end{tikzpicture}
\caption{\textbf{Perception in one RoboDojo episode}, put the bottles into the dustbin (outputs shortened). (a) \texttt{find} asks SAM~3 for ``bottle'' in the head picture and gets four outlines. (b) \texttt{measure} turns each outline into a Thing. The left wrist camera also sees bottle~3, so its position comes from two views (within 2\,mm). Only the head camera sees bottle~1, so the result names a wrist pose that adds a view. (c) From that pose, bottle~1 is measured in two views (within 3\,mm) and then grasped. The episode succeeded.}
\label{fig:perception}
\end{figure}

\texttt{measure} converts outlines into a 3D box. Without depth, as on RoboDojo, each mask defines a cone of rays; the object is also segmented in the other cameras and lies within the visual hull of their cones. The top and footprint of the hull, together with a rotated rectangle fitted to the footprint, give the centre, size and \texttt{grip\_yaw}, along with an error estimate. When only one view is available, the height cannot be measured; the result states this and proposes a wrist pose that would add a view (\cref{fig:perception}). Where the benchmark renders depth, the same calls back-project the mask's pixels instead. \texttt{to\_world} maps a single pixel to a world point, on a plane at a given height, from a second view or on the surface under the pixel. From the start images, low box-like objects are localized within a few millimetres; for round, hollow or tall objects, a wrist camera about 25\,cm above the object reduces the error to 1--2\,mm (\cref{app:perception}).

\subsection{Adapters, sandbox and prompt}
\label{sec:adapters}

The API is identical in every setting; below it, an adapter translates requests into the setting's own actions and reads its cameras and arm state. \Cref{tab:adapters} lists what lies below the API in each setting, with the lines of Python written for that setting alone and the length of its system prompt in characters. On the ARX X5 arms of RoboDojo and RoboTwin~2.0, HexaShell solves inverse kinematics itself; on the Franka Panda of LIBERO-PRO and RoboCasa365, it drives robosuite's operational-space controller with a feed-forward planner; and on the real robots, a separate service, URAI \cite{urai2026}, plans and executes every motion (\cref{app:motion,app:real}). The shared core comprises 5,577 lines of vision, 1,255 of robot API, 1,622 of agent loop and 806 of sandbox, plus 2,072 lines of motion for the two ARX X5 settings. The three benchmarks whose simulators run in a server process also share 4,134 lines (server, protocol and operational-space control). RoboTwin~2.0 has a second version of three prompt parts, for perception with depth.

\begin{table}[!htbp]
\centering
\caption{\textbf{Adapters and prompts.} The five settings share one core; only the adapter and the prompt are written for each.}
\label{tab:adapters}
\small
\setlength{\tabcolsep}{4pt}
\begin{tabularx}{\linewidth}{@{}lll lYrr@{}}
\toprule
Setting & Simulator & Robot & Cameras & Control & Code lines & Prompt chars. \\
\midrule
RoboDojo & Isaac Sim & Dual ARX X5 & \mbox{Head + wrists; RGB} & Joint targets; IK & 352 & 34,992 \\
RoboTwin~2.0 & SAPIEN & Dual ARX X5 & \mbox{Head + wrists; RGB} & Joint targets; IK & 1,449 & 27,208 \\
LIBERO-PRO & MuJoCo & Fixed Panda & Front + wrist; RGB-D & OSC & 694 & 27,062 \\
RoboCasa365 & MuJoCo & Mobile Panda & Base + wrist; RGB-D & OSC + base & 1,373 & 30,959 \\
Real robot & -- & Dual PiPER-X & \mbox{Head + wrists; RGB-D} & Joint targets; IK & 907 & 11,110 \\
\bottomrule
\end{tabularx}
\end{table}

\label{sec:sandbox}
Cells run in a separate worker process with one persistent Python namespace, and every \texttt{robot} call is forwarded to the host process, which owns the arms, cameras and perception. Landlock and seccomp confine the worker, so that the simulator, the task code, the skill sources and the network remain out of reach whatever a cell does (\cref{app:sandbox}).

\label{sec:prompt}
The system prompt is shared by all tasks of a benchmark and consists of six parts in a fixed order: the job and its rules, the world and the robot, the API, seeing and measuring, moving and handling objects, and checking results and recovering from errors. Their text is written for each robot and world, 27 to 35 thousand characters per simulated benchmark and 11 thousand on the real robot (\cref{app:prompt}). The task card is built from the first observation: the instruction, the step budget, the start pose of the arms, the start images and, on RoboDojo, the task's score table.

\subsection{Task skills}
\label{sec:skills}

A skill is a Markdown file for one task: about a page of guidance, not a program. Every skill on LIBERO-PRO, RoboCasa365 and RoboTwin~2.0 has five sections:
\begin{itemize}
    \item \textbf{Judging}: what counts as done, as earlier runs showed it, with no internal threshold of the checker;
    \item \textbf{Flow}: the steps that worked, in order, with the calls to use;
    \item \textbf{Verified parameters}: numbers measured in earlier runs, such as grasp heights, offsets and openings;
    \item \textbf{Common failures and fixes}: what went wrong and what to do instead;
    \item \textbf{Typical step costs}: how many steps each phase took, so that the agent can plan its budget.
\end{itemize}
RoboDojo skills also have a section of final checks and, for the twelve Generalization tasks, one on clutter. \Cref{app:skills} shows three skills in full.

Skills are drafted by an analysis agent, which reads the records of earlier runs of a group of related tasks, including every cell, output and image, and marks the evidence behind each claim; a person reviews every draft. A skill may contain only what the agent could observe: no positions read from the simulator, no internal names and no checker thresholds, a constraint enforced by a checking script. The median skill has 39 to 61 lines, depending on the benchmark, and only three of the 205 skills contain code (\cref{app:skill-details}).

\FloatBarrier
\section{Experiments}
\label{sec:experiments}

The experiments address three questions: how far the general tools go without any task-specific input (\cref{sec:main}); whether the harness transfers to other simulators and to real robots when only the adapter and the prompt change (\cref{sec:universal}); and what task skills add, measured both against runs without them and against a toolset written for each task (\cref{sec:knowledge}). \Cref{sec:setup} describes the benchmarks and the evaluation protocol, and \cref{sec:failures} analyses the failures that remain.

\subsection{Experimental setup}
\label{sec:setup}

We evaluate on four simulated benchmarks and on real robots. RoboDojo \cite{robodojo2025} has 42 tabletop tasks for two ARX X5 arms in Isaac Sim, in five dimensions: Generalization, Precision, Long-Horizon, Memory and Open. Its eight Open tasks receive no skills, so that they measure the general components alone. On RoboCasa365 \cite{nasiriany2026robocasa365} we run the 50 target tasks of its leaderboard, kitchen tasks for a Franka Panda on a wheeled base; on LIBERO-PRO \cite{zhou2025liberopro}, the four LIBERO suites under its position, object, language and task perturbations; and on RoboTwin~2.0 \cite{chen2025robotwin}, its 50 dual-arm tasks in randomized scenes with unseen instructions. Prompts, tools and skills are built on development episodes, and the test episodes differ in their layouts, seeds, kitchens, initial states or instruction templates. Every simulated run uses GPT-6-Astra, at reasoning effort xhigh for the RoboDojo and RoboCasa365 runs of \cref{tab:main,tab:robocasa}, except RoboDojo's Open tasks, and high elsewhere. RoboDojo reports a success rate (SR) and a score, both averaged over the tasks of each dimension and then over the dimensions; the other benchmarks report the share of successful episodes. All results are our own evaluations on public test episodes, with fewer episodes than the official protocols; \cref{app:protocol} gives the protocols and the choice of compared methods.

\subsection{Task-agnostic performance}
\label{sec:main}

\begin{table}[!htbp]
\centering
\caption{\textbf{RoboDojo results.} Score and success rate (SR, \%) overall, and success rate in each dimension. $^*$Our own evaluation on the public test layouts; the leaderboard entries are RoboDojo's official evaluation.}
\label{tab:main}
\small
\setlength{\tabcolsep}{5pt}
\begin{tabular}{@{}lccccccc@{}}
\toprule
Method & Score & SR & Gen. & Prec. & LH & Mem. & Open \\
\midrule
\multicolumn{8}{@{}l}{\emph{RoboDojo leaderboard}} \\
Q0-Preview \cite{qoo2026q0} & 74.20 & 68.87 & 58.67 & 58.50 & 58.50 & 93.67 & 75.00 \\
GO-3 Nano Preview \cite{agibot2026go3nano} & 61.90 & 56.42 & 33.67 & 76.00 & 53.50 & 63.67 & 55.25 \\
PyRUA-Lean \cite{si2026pyrua} & 55.26 & 50.35 & 43.17 & 38.25 & 43.75 & 68.33 & 58.25 \\
HexaShell-Preview (\cref{sec:v10}) & 53.18 & 46.85 & 34.50 & 32.00 & 31.50 & 81.00 & 55.25 \\
WorldScape-R$^2$SI \cite{manifold2026r2si} & 46.58 & 39.35 & 35.83 & 15.50 & 34.25 & 63.67 & 47.50 \\
EmbodiedEvo \cite{li2026embodiedevo} & 45.08 & 38.95 & 30.83 & 16.25 & 21.75 & 76.67 & 49.25 \\
WALL-OSS-1.0-lite \cite{xsquare2026wallosslite} & 42.59 & 34.97 & 37.94 & 29.08 & 32.25 & 56.00 & 19.58 \\
VPP2 & 39.26 & 32.26 & 29.50 & 35.17 & 37.00 & 55.56 & 4.08 \\
PhysicalRSI \cite{mmlab2026physicalrsi} & 36.27 & 31.38 & 15.62 & 32.50 & 37.33 & 46.56 & 24.92 \\
GPT-6-Astra, the RoboDojo team's agent \cite{zhang2026unexpected} & 28.97 & 22.48 & 30.50 & 4.00 & 8.25 & 38.67 & 31.00 \\
$\pi_{0.5}$ \cite{intelligence2025pi05} & 11.44 & 6.93 & -- & -- & -- & -- & -- \\
\midrule
HexaShell, no skills$^*$ & 75.73 & 69.92 & 64.44 & 52.50 & 54.58 & 98.89 & 79.17 \\
\textbf{HexaShell, with skills}$^*$ & \textbf{88.81} & \textbf{85.00} & 77.78 & 86.25 & 82.92 & 98.89 & 79.17 \\
\bottomrule
\end{tabular}
\end{table}

\begin{table}[!htbp]
\centering
\caption{\textbf{RoboCasa365 results.} Success rate (\%) on the 50 target tasks; the three groups of rows use different episodes and protocols.}
\label{tab:robocasa}
\small
\begin{tabular}{@{}lcccc@{}}
\toprule
Method & Atomic & Comp.\ seen & Comp.\ unseen & Overall \\
\midrule
\multicolumn{5}{@{}l}{\emph{Leaderboard}} \\
Awomo-0.5I \cite{awomo2026awomo05i} & 86.5 & 58.8 & 45.2 & 64.4 \\
minionx & 81.8 & 60.0 & 43.6 & 62.6 \\
Awomo-0.5V & 85.0 & 60.4 & 39.4 & 62.5 \\
Axiom-0 & 86.5 & 59.4 & 35.8 & 61.6 \\
Paimon-0 & 79.3 & 55.8 & 36.7 & 58.1 \\
$\pi_{0.5}$ \cite{intelligence2025pi05} & 39.6 & 7.1 & 1.2 & 16.9 \\
\midrule
\multicolumn{5}{@{}l}{\emph{Reported in papers}} \\
GPT-5.6-Sol, few-shot, as reported in \cite{xiaomi2026omnibrain0} & 81.1 & 45.0 & 27.5 & 52.4 \\
OmniBrain-0 RSI \cite{xiaomi2026omnibrain0} & 74.4 & 36.3 & 15.0 & 43.2 \\
PyRUA-Lean \cite{si2026pyrua} & 86.7 & \multicolumn{2}{c}{39.4 (both groups)} & -- \\
\midrule
HexaShell, no skills & 65.6 & 50.6 & 61.0 & 59.3 \\
\textbf{HexaShell, with skills} & \textbf{91.1} & \textbf{85.0} & \textbf{80.0} & \textbf{85.6} \\
\bottomrule
\end{tabular}
\end{table}

\Cref{tab:main} compares two runs of HexaShell, which differ only in the skills, with the RoboDojo leaderboard. The leaderboard gives RoboDojo's official evaluation of 30 October 2026, run by the benchmark's team with 50 episodes per task and seed on layouts that are not public. Our runs use 30 test episodes per task (seeds 0--2, layouts 10--19), and we report the mean over the three seeds, whose standard deviation is below one point.

Without any task-specific input, HexaShell reaches a success rate of 69.9 and a score of 75.7 on RoboDojo, above every entry of its leaderboard; the strongest, Q0-Preview, reaches 68.9 and 74.2. The contribution of the harness is best isolated with the model held fixed. Run as an agent by the RoboDojo team, GPT-6-Astra reaches 22.5 \cite{zhang2026unexpected}, and RoboDawn reports 35.7 zero-shot and 47.2 one-shot with the same model \cite{guo2026robodawn}. With the model unchanged, the tools, prompt and loop of HexaShell raise the success rate from 22.5 to 69.9.

\label{sec:tool-use}
The agent operates almost entirely through the API. In both RoboDojo runs, \texttt{goto} accounts for 41--43\% of all robot calls, \texttt{look} 19--21\%, \texttt{measure} 10--11\%, \texttt{to\_world} 8--10\%, \texttt{gripper} 7--8\% and \texttt{find} 2--3\%. The median episode measures 16 times without skills and 11 times with them, typically again after a hand has moved something, or from a wrist camera when a result warns that a single view could not determine the height. Only about one episode in ten processes images in its own code, more than half of them on the two tasks with T-shaped blocks (\cref{app:tool-use}). The perception calls thus cover the needs of nine episodes in ten, and for the rest the sandbox still lets the agent write its own code.

\subsection{Transfer across simulators and real robots}
\label{sec:universal}

The same API, loop and sandbox run on three simulators, two kinds of simulated arm on fixed and wheeled bases, cameras with and without depth, and real robots; only the adapters and the prompt texts differ (\cref{tab:adapters}). With skills, HexaShell exceeds the best published result we are aware of on every simulated benchmark (\cref{fig:teaser,tab:main,tab:robocasa,tab:libero,tab:robotwin}).

On RoboCasa365 it reaches 59.3\% without skills and 85.6\% with them, against 64.4\% for the best leaderboard entry, Awomo-0.5I (\cref{tab:robocasa}). The margin is largest on the composite tasks: 85.0\% against 58.8\% on the seen ones and 80.0\% against 45.2\% on the unseen. On the same 1,500 episodes the skills gain 466 and lose 72 ($p<0.001$), most of them on tasks at the sink and on appliance doors and buttons. On LIBERO-PRO it reaches 78.3\% without skills and 96.7\% with them, and on 54 tasks a paired control raises it from 252 to 531 of 540 episodes; other coding agents report 94.0 to 96.5\% on subsets of the benchmark \cite{wang2026harnesspai,lu2026aspire,si2026pyrua}, and its leaderboard, which lists only VLAs, tops out at 53\%. On RoboTwin~2.0, in randomized scenes with unseen instructions, it reaches 82.2\% without skills on the learning seeds and 96.8\% with them on the test, against 81.5\% for the top leaderboard entry, GO-3 Nano Preview, and 90.4\% for the best result in a paper.

\begin{figure}[!t]
\centering
\newcommand{\realrow}[3]{{\raggedright\sffamily\footnotesize\textbf{#1}\enspace #2\par}\vspace{1.5pt}\includegraphics[width=\linewidth]{fig/img/real_#3.jpg}\par\vspace{4pt}}
\realrow{Shuffle blocks}{10/10}{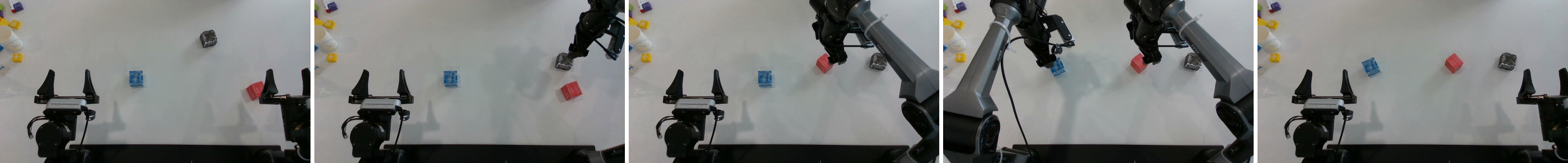}
\realrow{Sort blocks by language instruction}{10/10}{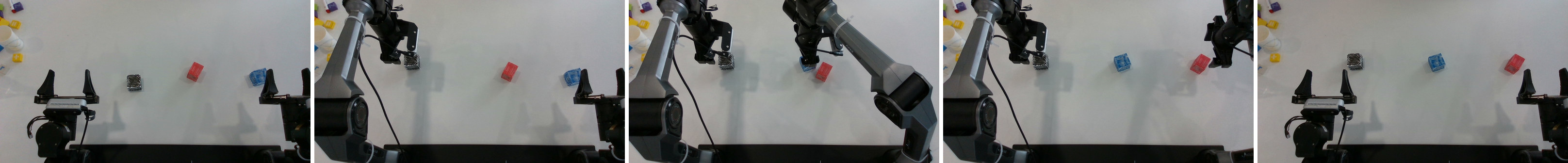}
\realrow{Stack blocks}{9/10}{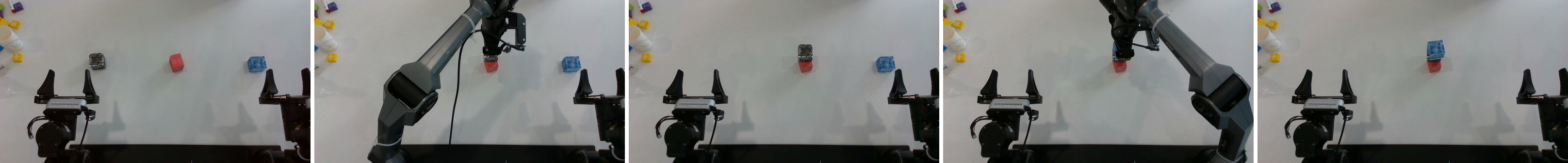}
\realrow{Unstack blocks}{8/10}{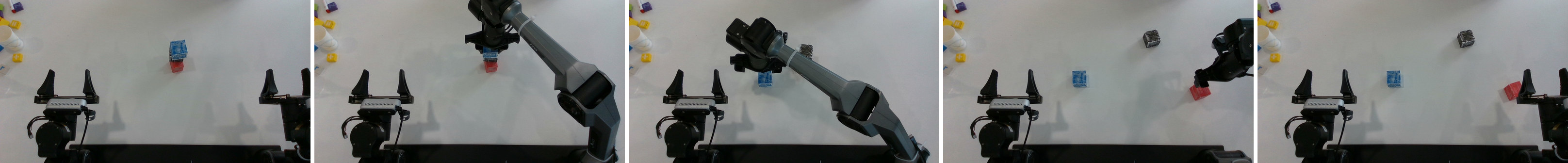}
\realrow{Cover blocks}{10/10}{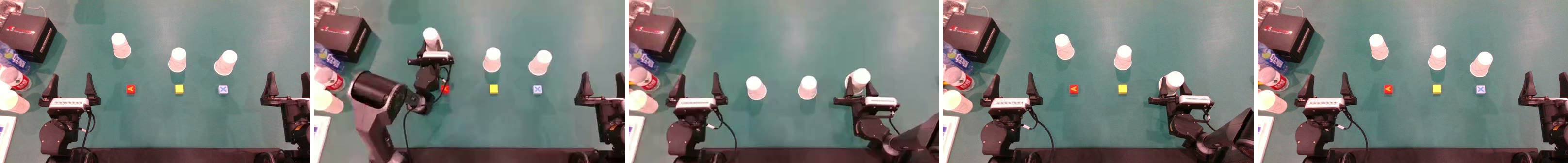}
\realrow{Sort pens and plush toys}{7/10}{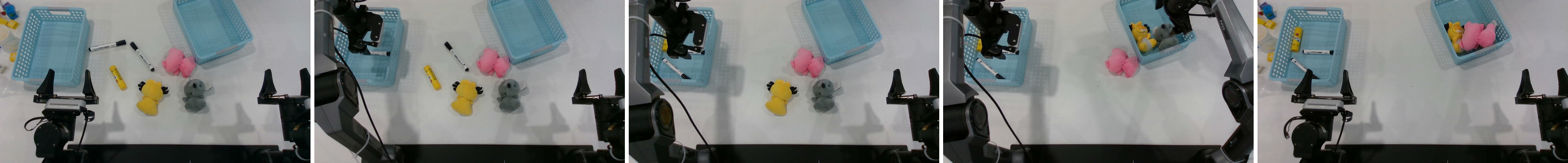}
\realrow{Put papaya in plate}{9/10}{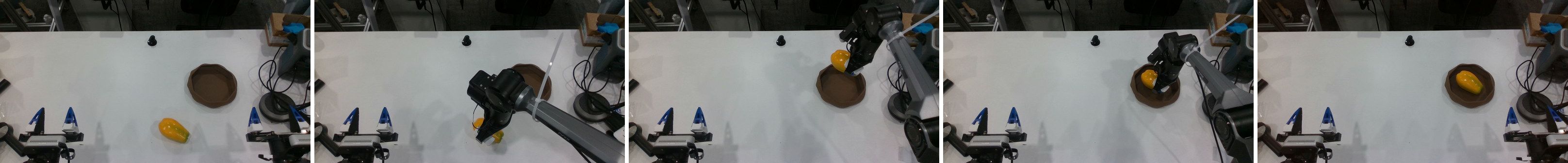}
\realrow{Take papaya out of plate}{5/10}{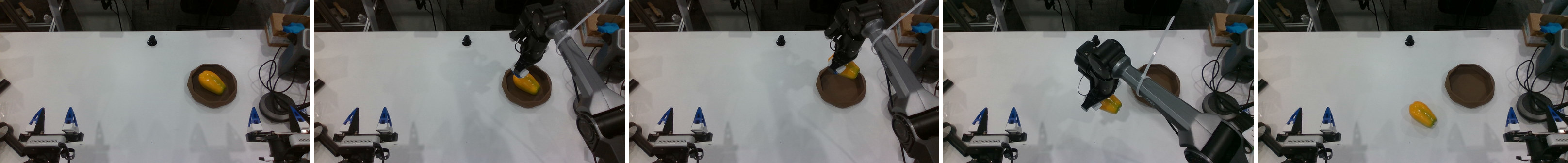}
\realrow{PushT}{8/10}{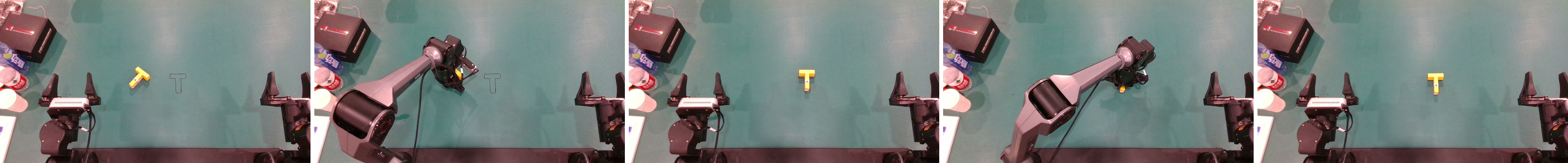}
\caption{\textbf{HexaShell on dual-arm PiPER-X robots}, seen from the head camera. One successful episode of each task in \cref{tab:real}, from its start on the left to its end on the right.}
\label{fig:real}
\end{figure}

\label{sec:real}
The same loop, sandbox and API also drive real robots: three dual-arm PiPER-X robots with RGB-D head and wrist cameras. The 907-line adapter passes each \texttt{goto} to URAI, which plans it, checks it for collisions and executes it; a refused plan is reported as an unreachable pose, as in simulation. The real robots have no segmentation service yet, so the model reads pixel coordinates off the images and turns them into points with \texttt{to\_world} on the measured depth. Over the nine tasks of \cref{tab:real}, HexaShell succeeds in 76 of 90 episodes. GPT-6.1-Terra writes the code for all nine and, in the absence of a checker, GPT-6-Astra judges each episode. The tasks include shuffling, stacking and unstacking blocks, sorting them by a language instruction, sorting pens from plush toys, putting a papaya in a plate and taking it out, and PushT. In cover blocks, the agent notes the colours of three blocks, covers them with cups from left to right, then uncovers them, and succeeds in 10 of 10. \Cref{fig:real} shows one successful episode of each task.

\begin{table}[!htbp]
\centering
\captionsetup{justification=raggedright,singlelinecheck=false}
\begin{minipage}[t]{0.37\linewidth}
\centering
\caption{\textbf{Real-robot results on dual-arm PiPER-X robots.}}
\label{tab:real}
\small
\setlength{\tabcolsep}{5pt}
\begin{tabular}{@{}lr@{}}
\toprule
Task & Successes \\
\midrule
Shuffle blocks & 10/10 \\
Sort blocks by language instruction & 10/10 \\
Stack blocks & 9/10 \\
Unstack blocks & 8/10 \\
Cover blocks & 10/10 \\
Sort pens and plush toys & 7/10 \\
Put papaya in plate & 9/10 \\
Take papaya out of plate & 5/10 \\
PushT & 8/10 \\
\bottomrule
\end{tabular}
\end{minipage}\hfill
\begin{minipage}[t]{0.60\linewidth}
\centering
\caption{\textbf{Task-specific tools versus shared tools.} RoboDojo success rate (SR) and score (\%).}
\label{tab:v10}
\small
\setlength{\tabcolsep}{3.5pt}
\renewcommand{\arraystretch}{1.3}
\begin{tabular}{@{}llrrrrrr@{}}
\toprule
System & Metric & Overall & Gen. & Prec. & LH & Mem. & Open \\
\midrule
HexaShell-TaskTools & SR & 46.85 & 34.50 & 32.00 & 31.50 & 81.00 & 55.25 \\
 & Score & 53.18 & 39.22 & 41.25 & 42.91 & 84.53 & 57.97 \\
\addlinespace[1em]
HexaShell, no skills & SR & 69.92 & 64.44 & 52.50 & 54.58 & 98.89 & 79.17 \\
 & Score & 75.73 & 69.86 & 60.38 & 67.02 & 98.89 & 82.50 \\
\addlinespace[1em]
HexaShell, skills & SR & \textbf{85.00} & \textbf{77.78} & \textbf{86.25} & \textbf{82.92} & \textbf{98.89} & \textbf{79.17} \\
 & Score & \textbf{88.81} & \textbf{82.22} & \textbf{88.25} & \textbf{92.21} & \textbf{98.89} & \textbf{82.50} \\
\bottomrule
\end{tabular}

\end{minipage}
\end{table}

\subsection{Effect of task skills}
\label{sec:knowledge}
\label{sec:v10}

Knowledge that a general harness lacks about a task can be supplied either as brief guidance, as in HexaShell, or as an elaborate task-specific toolset. HexaShell-TaskTools, an earlier version of our agent listed as HexaShell-Preview on the RoboDojo leaderboard, takes the latter approach. An acting agent (the Codex command-line agent with GPT-6-Astra) drives the robot through a command line. After each failed episode, an optimizer agent of the same model writes or repairs the task's tools, with their tests, and the same layout is tried again. Each task was developed for up to 12 hours on the first ten layouts of seed~0, the same development layouts that HexaShell used. The 769 optimizer rounds left 327 task commands in about 78,000 lines of Python, which were then frozen. In RoboDojo's official evaluation, on a hidden seed with 50 episodes per task, they reach a success rate of 46.9 and a score of 53.2. \Cref{tab:v10} sets this official result against our two test runs of \cref{tab:main}, overall and per dimension, with the Open tasks counted as in \cref{tab:main}.

The resulting code is large and generalizes poorly. Re-run after freezing, the tools pass only 152 of the 420 development episodes on which they were built (36\%). An audit further found object sizes copied from the scenes into the tools of seven tasks, a shortcut that a measurement within the episode should replace. HexaShell, by contrast, keeps no tools for any task. Its 34 RoboDojo skills are plain text, about 2,700 lines in total or about a page per task, each of which a person can read and verify; only three contain code (\cref{app:skill-details}).

Brief guidance proves more effective than a large toolset. Even without skills, HexaShell outperforms HexaShell-TaskTools in every dimension. With skills, it reaches a success rate of 85.0 and a score of 88.8, and the gap is largest precisely where per-task code was expected to help most: Precision and Long-Horizon, 86.3 and 82.9 against 32.0 and 31.5. Although the official evaluation and our test runs use different episodes, the ranking also holds on public episodes: on 420 episodes of seed~0, with the same layouts outside Generalization, the frozen tools of HexaShell-TaskTools pass 157, HexaShell 284 without skills and 351 with them. Since the two versions also differ in the loop, perception, motion and prompt, this comparison establishes that general tools with short skills outperform tools written for each task, but not which change matters most. A paired comparison isolates the skills: of the 1,020 RoboDojo episodes outside Open, 234 succeed only with skills and 37 only without ($p<0.001$; \cref{sec:ablation}).

\begin{figure}[t]
\centering
\includegraphics[width=\linewidth]{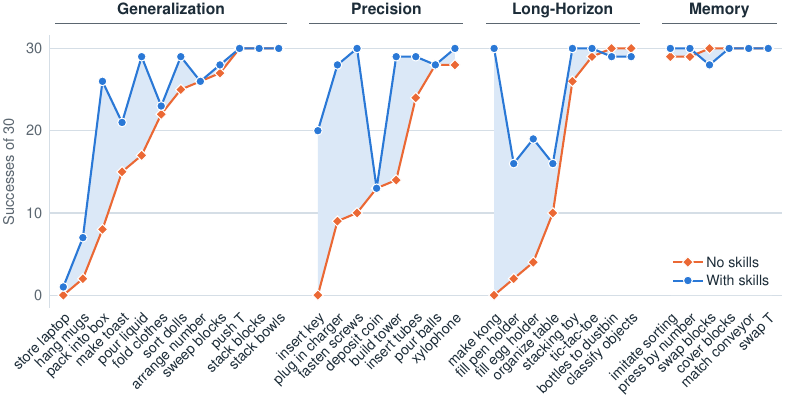}
\caption{\textbf{RoboDojo per task.} Successes out of 30 test episodes without skills (diamonds) and with skills (dots), sorted within each dimension by successes without skills. Open tasks have no skills and are omitted. All numbers are given in \cref{tab:pertask-robodojo}.}
\label{fig:pertask}
\end{figure}

\label{sec:skill-analysis}
The gains from skills concentrate on tasks that the general components could not solve (\cref{fig:pertask}): make kong rises from 0 to 30 of 30, fasten screws from 10 to 30, plug in charger from 9 to 28, pack objects into box from 8 to 26 and build tower from 14 to 29, while no task loses more than two episodes. These tasks also failed differently from the rest: without a skill, 37 of their 109 failures involved a slip and 45 involved neither a slip nor a blocked motion, against 171 and 31 of the 244 failures of the other tasks outside Open. In these tasks the agent could perceive, grasp and carry the objects; what it lacked was task knowledge of three kinds.
\begin{itemize}
    \item \emph{Success conditions that the task card leaves open.} In make kong, the tile to place is the one drawn from the wall, not the opponent's discard: ten development runs that moved the discard all scored 0. In pack objects into box, the card credits items placed in ``the aligned box'' without saying when a box counts as aligned; the skill records that a shifted but upright box still counted, and that gripping, hooking or dragging a moved box restored it in none of 13 development runs. Without the skill, the agent gripped the box in all 22 failed test episodes, in 17 of them before handling any item.
    \item \emph{Parameters that only contact reveals.} The fasten-screws skill says that the ``screws'' are bolts fixed to the table, so the agent carries each nut to its bolt and never picks a bolt up, and that a pushed turn down to a grasp height of 0.793--0.800\,m seated 8 of 11 first placements. The plug-in-charger skill below is another example.
    \item \emph{The ordering of steps.} The build-tower skill says to build each level from one pillar of each side of the table. The three development runs that placed both lower pillars with one arm never got past the lowest score.
\end{itemize}

\begin{figure}[t]
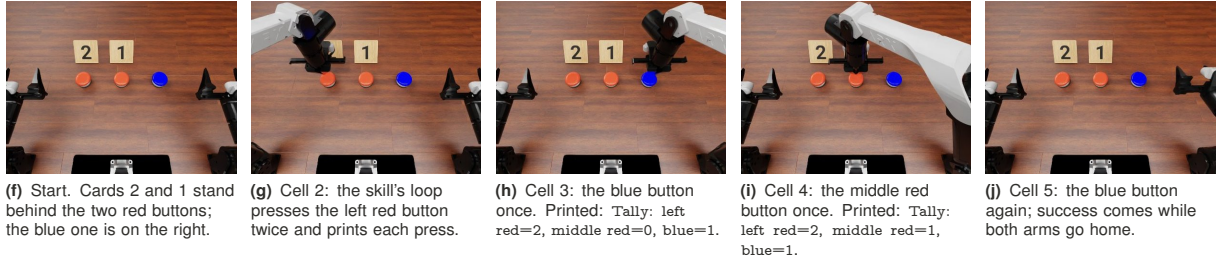

\centering
\begin{tikzpicture}[
  x=1mm, y=1mm, font=\sffamily\fontsize{6.2}{7.4}\selectfont, text=rsink,
  row/.style={font=\sffamily\bfseries\fontsize{7.5}{9}\selectfont, anchor=base west, inner sep=0pt},
  img/.style={anchor=north west, inner sep=0pt},
  txt/.style={anchor=north west, inner sep=0pt, align=left, text width=30.4mm},
]
\newcommand{\pl}[1]{\textbf{(#1)}\enspace}
\node[row] at (0,0) {Plug in charger\textmd{\enspace test seed 1, layout 12, effort xhigh: success at step 179 of 400}};
\foreach \f/\t [count=\i from 0] in {
  a/{\pl{a}Start. Dry runs report both arms not reachable at the strip at the insertion yaw, $-27.6^\circ$.},
  b/{\pl{b}Cells 4--6: the left arm moves the strip 37\,cm, to where the right arm can insert (77 steps).},
  c/{\pl{c}Cells 8--9: left wrist, $\times$2. \texttt{to\_world} turns the slot's pixel into a point on the strip's top, about 0.6\,mm per pixel.},
  d/{\pl{d}Cell 13: the level right hand takes the prongs down to the seat, 0.800\,m, with \texttt{contact=True} (head, $\times$4).},
  e/{\pl{e}Cells 14--15: it lets go at the seat; success comes while both arms go home.}} {
  \node[img] (c\i) at (32.4*\i,-2.4) {\includegraphics[width=30.4mm]{fig/img/charger_\f.jpg}};
  \node[txt] at ($(c\i.south west)+(0,-1.2)$) {\t};
}
\node[row] at (0,-42.2) {Press by number\textmd{\enspace test seed 0, layout 14, effort high: success at step 109 of 700}};
\foreach \f/\t [count=\i from 0] in {
  a/{\pl{f}Start. Cards 2 and 1 stand behind the two red buttons; the blue one is on the right.},
  b/{\pl{g}Cell 2: the skill's loop presses the left red button twice and prints each press.},
  c/{\pl{h}Cell 3: the blue button once. Printed: \texttt{Tally: left red=2, middle red=0, blue=1}.},
  d/{\pl{i}Cell 4: the middle red button once. Printed: \texttt{Tally: left red=2, middle red=1, blue=1}.},
  e/{\pl{j}Cell 5: the blue button again; success comes while both arms go home.}} {
  \node[img] (p\i) at (32.4*\i,-44.6) {\includegraphics[width=30.4mm]{fig/img/press_\f.jpg}};
  \node[txt] at ($(p\i.south west)+(0,-1.2)$) {\t};
}
\end{tikzpicture}
\caption{\textbf{Two RoboDojo episodes with skills} (\cref{sec:cases}). Top: insertion. Dry runs show that no arm can insert level where the strip lies, so the other arm first moves the strip. Bottom: pressing buttons by count (\cref{app:episode}). Each cell presses one button as often as needed and prints the count; the episode takes five cells.}
\label{fig:cases}
\end{figure}

\label{sec:cases}
In plug in charger, the agent must insert the two prongs of a charger lying on the table into a power strip. Without a skill it found, grasped and carried the charger, yet passed only 9 of 30 episodes; in its own words, the charger ``did not remain seated and tipped onto the table beside the strip''. The skill, 61 lines without code, states what the development runs had shown: insert with a level hand at the strip's grasp yaw $\pm90^\circ$; seat the grasp point 1.8--2\,cm above the strip's measured top, asking for 0.4\,cm lower because a held load stops short; let go only at the seat (0 of 8 counted otherwise); and, since a charger tilted by $3^\circ$ or more never went in (0 of 13), check both arms with \texttt{dry\_run} and let the other arm move the strip if neither can insert level. With the skill the task passes 28 of 30 (\cref{fig:cases}, top). None of these facts is visible in the images or stated on the card. Press by number (\cref{fig:cases}, bottom) illustrates a different kind of fix: a counting rule that keeps the agent from pressing again to make sure (\cref{app:press}). \Cref{app:cases} follows six more tasks through one test episode each, without and with the skill. Skills also make episodes cheaper, reducing the turns by 17\%, the steps by 19\% and the reasoning tokens by 44\% (\cref{sec:cost}).

\subsection{Failure analysis}
\label{sec:failures}

The agent ends most failed episodes itself, mostly because the remaining steps cannot pay for another attempt, and most failures still earn part of the task's score (\cref{app:failures}). The most frequent cause of failure is an object slipping from the fingers. With skills, 153 of the 1,020 RoboDojo episodes outside Open contained a slip, and 46\% of them succeeded, against 91\% of the others. Skills make slips rarer but do not eliminate them. The Open task store tools in toolbox passes no episode: as the arm lifts, a long tool gripped near its middle rotates between the fingertips and slides out, and scripted tests outside the agent never placed the pliers either (\cref{app:toolbox}). These failures originate in the grasps and motions below the API, which no skill can change; better grasping at that level would benefit every task at once.

\FloatBarrier
\section{Related Work}
\label{sec:related}

\paragraph{Code generation for robot control.} Code as Policies \cite{liang2023code} has a language model write programs over perception and control primitives. ProgPrompt \cite{singh2022progprompt} and Inner Monologue \cite{huang2022inner} plan with a language model and feedback from the scene, and SayCan \cite{ahn2022saycan} grounds language in a fixed library of skills. VoxPoser \cite{huang2023voxposer} composes value maps for a motion planner, while ReKep \cite{huang2024rekep} and MOKA \cite{liu2024moka} let a vision model mark keypoints that code then turns into motion. RoboCodeX \cite{mu2024robocodex} and RoboScript \cite{chen2024roboscript} generate manipulation code from images and text, and language models also write reward functions for policy learning \cite{yu2023language,ma2023eureka}. In these systems the primitives are written by hand, and the model composes them into a program for each instruction. HexaShell likewise relies on general, hand-written primitives, but it operates in a closed loop: the model writes one short cell at a time and chooses the next step only after reading the printed results and the images.

\paragraph{VLM agents as robot policies.} A frontier VLM can now serve as the policy itself. Zhang \etal~\cite{zhang2026unexpected} evaluate GPT-6-Astra on RoboDojo through a fixed end-effector harness, with no learned policy in the loop. Show-Harness \cite{chen2026showharness}, Agent as Policy \cite{jia2026agentaspolicy}, VIA \cite{hu2026via} and World Action Agent \cite{zhang2026waa} drive robots through a VLM agent and a fixed interface. Robo-Harness K1 adds perception tools for such agents \cite{li2026roboharnessk1}, and RoboDawn reports GPT-6-Astra on RoboDojo zero-shot and one-shot \cite{guo2026robodawn}. Harness VLA \cite{zhang2026harnessvla} lets a memory-guided agent steer a frozen VLA through manipulation primitives, and OmniBrain-0 \cite{xiaomi2026omnibrain0} is a single agentic model for embodied tasks. CaP-X \cite{fu2026capx} and LIBERO-Agent \cite{diao2026liberoagent} benchmark coding agents and general agents on manipulation, RLE-Bench \cite{ma2026rlebench} tests coding agents as robot learning engineers, and EmbodiedLab \cite{gao2026embodiedlab} tests whether agents act on a scene to find out what a task requires. The agent track of the RoboDojo leaderboard lists Q0-Preview \cite{qoo2026q0}, PyRUA-Lean \cite{si2026pyrua} and EmbodiedEvo \cite{li2026embodiedevo} as standalone agents, and GO-3 Nano Preview \cite{agibot2026go3nano}, WorldScape-R$^2$SI \cite{manifold2026r2si} and PhysicalRSI \cite{mmlab2026physicalrsi} as agents that direct a learned model (\cref{app:comparisons}). HexaShell belongs to this family and differs mainly in its interface, whose calls return measured 3D positions with error estimates, check a pose before it consumes steps, and report problems first, together with what to change.

\paragraph{Self-improving coding agents.} Coding agents read a repository, run its tests and fix the code \cite{chen2021codex,jimenez2024swebench,yang2024sweagent,wang2025openhands}, and Reflexion \cite{shinn2023reflexion} keeps verbal lessons from failed attempts. Voyager \cite{wang2023voyager} grows a library of code skills while playing Minecraft. Toolformer \cite{schick2023toolformer} learns when to call tools, LATM \cite{cai2024latm} and CRAFT \cite{yuan2024craft} have language models write and retrieve tools, and self-improving systems modify their own programs and retain the changes that raise a score \cite{zhang2025dgm,novikov2025alphaevolve,lu2024aiscientist}. In robotics, recent harnesses let coding agents program, test and refine robot skills \cite{elmaaroufi2026rho,lu2026aspire,xiao2026enpire,wang2026harness,ding2026zetta,talwar2026local,chen2026gap,zhang2026rats}, some starting from a few human demonstrations \cite{liu2026rapid,kang2026encore}. URAI pairs a programming agent that writes motion tools with an execution agent that calls them \cite{urai2026}. EmbodiedEvo turns physical feedback into code skills and diagnostic lessons \cite{li2026embodiedevo}, MEMENTO evolves code policies with a memory of past programs \cite{sygkounas2026memento}, and Gao \etal~\cite{gao2026selfevolving} treat tools, traces and memory as code that a physical coding agent revises. In most of these systems, what is retained for a task is code, as in HexaShell-TaskTools, an earlier version of our agent that wrote and repaired tools for each task (\cref{sec:v10}). In HexaShell the code is general and fixed, and what is retained for a task is only a short, reviewed skill in natural language. In this respect it is closer to Reflexion's verbal lessons than to a skill library.

\paragraph{Visual grounding for manipulation.} SAM \cite{kirillov2023sam}, SAM~2 \cite{ravi2024sam2} and SAM~3 \cite{carion2025sam3} segment objects from points and boxes, and SAM~3 also from short phrases. VLMs, by contrast, judge positions and distances in an image poorly \cite{fu2024blink,rahmanzadehgervi2024blind}; SpatialVLM \cite{chen2024spatialvlm} and SpatialRGPT \cite{cheng2024spatialrgpt} train them on spatial data to close this gap. Monocular depth networks predict depth from a single image \cite{yang2024depthv2,bochkovskii2025depthpro,wang2025moge,wang2025moge2}. Multiple-view geometry recovers a point from two calibrated views \cite{hartley2004multiple}, plane sweep tests candidate depths by comparing views \cite{collins1996space}, and both are standard in OpenCV \cite{bradski2000opencv}. Active perception moves the sensor to obtain a better view \cite{bajcsy1988active,aloimonos1988active}. In HexaShell the VLM never estimates a distance itself: it names an object or points at pixels, SAM~3 returns the outline, and calibrated geometry or rendered depth yields the position with an error estimate. The model decides when a wrist camera should add a view, a simple form of active perception.

\paragraph{Learned policies and benchmarks.} Vision-language-action models \cite{brohan2023rt2,kim2024openvla,octo2024,black2024pi0,intelligence2025pi05,bjorck2025groot,zhu2026actionunet} learn manipulation from demonstrations, some collected with low-cost hardware \cite{zhao2023aloha}. They are evaluated on suites such as LIBERO \cite{liu2023libero}, RoboCasa \cite{nasiriany2024robocasa,nasiriany2026robocasa365}, RoboTwin~2.0 \cite{chen2025robotwin} and SIMPLER \cite{li2024simpler}, and in autonomous real-world evaluation \cite{zhou2025autoeval}. LIBERO-PRO \cite{zhou2025liberopro} and LIBERO-Plus \cite{fei2025liberoplus} perturb LIBERO's objects, positions, instructions and scenes, so that a policy cannot succeed by memorizing the training layouts. RoboDojo \cite{robodojo2025} fixes a step budget per task, tests on several seeds and pairs clean layouts with cluttered ones. Most leaderboard entries on these benchmarks are learned policies, so our comparisons set an agent that uses no demonstrations against policies trained on many.

\FloatBarrier
\section{Limitations}
\label{sec:limitations}

Our results come from our own evaluations on public test episodes, with fewer episodes than the official protocols. A skill requires runs of its task, so a new task starts at the level of the general components, and we have not measured how few runs suffice. We have not yet quantified the contribution of SAM~3 and the geometry by switching them off. Slips of held objects remain the most common failure, and the real robots run without SAM~3, with few episodes so far.

\FloatBarrier
\section{Conclusion}
\label{sec:conclusion}

We presented HexaShell, a harness in which a VLM controls robots by writing one Python cell per turn against a small robot API. Task-agnostic tools carry the agent most of the way: SAM~3 for segmentation and NumPy and OpenCV geometry for localization, shared unchanged by four simulated benchmarks. The real robots use the same loop, sandbox and geometry but have no SAM~3 yet, so there the model picks the pixels itself. About a page of guidance per task, distilled from earlier runs, supplies what such tools cannot know, namely the success condition, parameters that only contact reveals and the order of steps, and in our own evaluations it lifts the agent above the best published result we are aware of on all four benchmarks.

Most remaining failures are slips of held objects and final placements that must be exact. Better grasps and motions below the API would benefit every task at once. They are the focus of our next steps, together with the official evaluations, more real-robot episodes and an ablation without SAM~3.

\clearpage

\bibliographystyle{plainnat}
\bibliography{main}

\clearpage
\appendix
\crefalias{section}{appendix}\crefalias{subsection}{subappendix}\crefalias{subsubsection}{subsubappendix}%
\FloatBarrier
\section{Evaluation protocol}
\label{app:protocol}

\begin{table}[!htbp]
\centering
\caption{\textbf{Evaluation protocols.} Development and test episodes, limits per episode and the official test protocol of each benchmark.}
\label{tab:protocol}
\small
\setlength{\tabcolsep}{4pt}
\begin{tabularx}{\linewidth}{@{}lYYYY@{}}
\toprule
 & RoboDojo & RoboCasa365 & LIBERO-PRO & RoboTwin~2.0 \\
\midrule
Tasks / skills & 42 / 34 & 50 / 50 & 156 / 71 & 50 / 50 \\
Development & Seed 0; layouts 0--9 & Seeds 0--19 & States 0--19 & 20 seeds/task; seen instructions \\
Test & Seeds 0--2; layouts 10--19 & Seeds 20--49 & States 20--29 & 15 expert-valid seeds/task; unseen instructions \\
Episodes & 1,260 & 1,500 & 1,560 & 750 \\
\addlinespace
Step limit & Per task & Per task & 220--520 & 400--1,700 \\
Reply limit & 100 & 40 / 100 & 40 & 40 \\
Time limit & 90 min & 120 min & 120 min & 120 min \\
Metric & Score; macro SR & SR & SR & SR \\
\addlinespace
Official test & Hidden layouts; 50/task/seed & 50/task & 50/task; 5 perturbations & 100/task/setting \\
\bottomrule
\end{tabularx}
\end{table}

\subsection{Benchmarks and runs}

\Cref{tab:protocol} lists, for each benchmark, the development and test episodes, the agent's limits per episode (\cref{app:loop}) and the official test protocol, which we did not run. On RoboCasa365 the reply limit is 100 on composite tasks and 40 on atomic ones, and LIBERO-PRO's step limits leave out its ten settling steps. The official evaluation of RoboTwin~2.0 covers clean as well as randomized scenes, and that of RoboCasa365 uses its pretraining kitchens. RoboDojo \cite{robodojo2025} has 42 tabletop tasks for two ARX X5 arms in Isaac Sim, in five dimensions: Generalization (12 tasks, half of whose test episodes add clutter), Precision (8), Long-Horizon (8), Memory (6) and Open (8). RoboCasa365 \cite{nasiriany2026robocasa365} has kitchen tasks for a Franka Panda on a wheeled base, and we run the 50 target tasks of its leaderboard. LIBERO-PRO \cite{zhou2025liberopro} perturbs the four LIBERO suites \cite{liu2023libero}: objects are moved (position), changed in appearance (object), the instruction is reworded (language), or a different task is asked in the same scene (task). We run these four perturbations. The fifth, which changes the environment, is not in the release and has to be generated; we have not run it. RoboTwin~2.0 \cite{chen2025robotwin} has 50 tasks for two ARX X5 arms in SAPIEN. We run its clean-to-randomized setting (``hard'' on its leaderboard): randomized scenes with clutter, textures, lighting and table height, and instructions from the templates kept for testing (unseen). The test seeds are chosen as in the official evaluation, the first seeds from 100000 whose scenes the benchmark's expert solves.

Every simulated run uses GPT-6-Astra through an OpenAI-compatible gateway (the real robots' models are given in \cref{app:real}). The two runs of \cref{tab:main} and the two of \cref{tab:robocasa} use reasoning effort xhigh, with at most 32,000 output tokens per reply on RoboDojo and 8,192 on RoboCasa365. All other runs use effort high with at most 8,192: the LIBERO-PRO and RoboTwin~2.0 results, the harness comparisons of \cref{tab:ablation}, and the RoboDojo Open tasks, whose one run both rows of \cref{tab:main} share. In that run, store tools in toolbox, which had passed none of its 30 episodes in an earlier run, was not run again and counts as 0. Two runs kept a larger picture window than that of \cref{app:loop}: that Open run (90 and 60 pictures and 10\,MB instead of 20, 14 and 8\,MB) and the RoboCasa365 run without skills (50 and 30 pictures).

Prompts, tools and skills were built on the development episodes of \cref{tab:protocol}, not on the test episodes. On RoboDojo these were five rounds on seed~0 and layouts 0--9 (R0--R4); on RoboCasa365, learning seeds 0--9 and validation seeds 10--19 of the pretrain50 split; on LIBERO-PRO, learning states 0--9 and validation states 10--19; and on RoboTwin~2.0, ten learning and ten validation seeds per task below 100000. Between R0 and R1, which added the first skills and a new motion planner, successes on the 34 tasks outside Open rose from 173 to 251 of 340, and R4 solved 330 of 422 episodes of all 42 tasks. The test episodes differ from the development ones in their layouts and seeds (RoboDojo), seeds and kitchens (RoboCasa365), initial states (LIBERO-PRO), and seeds and instruction templates (RoboTwin~2.0). RoboDojo's layouts 10--19 are public, while its official evaluation uses layouts that are not. HexaShell has been submitted to RoboDojo's official evaluation.

\subsection{Metrics and comparisons}
\label{app:comparisons}

RoboDojo reports a score, the task progress that each task's score table defines, and a success rate (SR). As RoboDojo computes them, both are averaged first over the tasks of each dimension and then over the five dimensions. The eight Open tasks therefore weigh as much as the twelve Generalization tasks, and the SR differs from the share of successful episodes. Without skills it is 69.92 for 857 of 1,260 episodes (68.0\%), and with skills 85.00 for 1,054 of 1,260 (83.7\%). We report the mean and standard deviation over the three seeds. The other benchmarks report the share of successful episodes, and RoboCasa365's overall number is the mean over its 50 tasks. A paired comparison runs the same episodes twice, and we test it with an exact two-sided sign test on the episodes that succeeded in only one of the two runs.

For each benchmark we list the leaderboard entries on the date given in its table and results reported in papers; we ran no other method. HexaShell-TaskTools's results come from RoboDojo's official evaluation and from our retests of its frozen tools (\cref{sec:v10}). Leaderboard entries follow the official protocol and papers use their own episodes, often on a subset of tasks or perturbations, so neither comparison with our evaluations is exact. \Cref{fig:teaser} draws the top three other entries of each leaderboard in its own ranking: on RoboDojo \cite{robodojoleaderboard}, Q0-Preview, GO-3 Nano Preview and PyRUA-Lean, by average score on 30 October 2026; on RoboCasa365 \cite{robocasa365leaderboard}, Awomo-0.5I, minionx and Axiom-0, leaving out Awomo-0.5V, a second model of the Awomo team; and on RoboTwin~2.0 \cite{robotwin2leaderboard}, GO-3 Nano Preview, PrismWAM* and PICA-WAM, by the mean of the easy and hard settings, drawn in the hard setting that we ran. The LIBERO-PRO leaderboard lists only VLAs, the best at 53\% over five perturbations, so there the figure draws the three highest results in papers, each on its own cells: HarnessPAI, 96.5\% on the position and task perturbations of three suites \cite{wang2026harnesspai}; ASPIRE with GPT-6-Astra, 96.0\% on the same perturbations of all four suites (our mean of its eight per-cell results; 98.8\% with Claude Opus 5.5) \cite{lu2026aspire}; and PyRUA-Lean with GPT-6-Astra, 94.0\% on 40 tasks with five seeds each \cite{si2026pyrua}.

RoboDojo's leaderboard \cite{robodojoleaderboard} marks each entry of its agent track as an agent alone or as an agent that directs a learned model. Q0-Preview \cite{qoo2026q0}, PyRUA-Lean \cite{si2026pyrua}, EmbodiedEvo \cite{li2026embodiedevo}, the RoboDojo team's GPT-6-Astra agent \cite{zhang2026unexpected} and HexaShell-Preview, our HexaShell-TaskTools (\cref{sec:v10}), are agents alone. GO-3 Nano Preview \cite{agibot2026go3nano} and WorldScape-R$^2$SI \cite{manifold2026r2si} direct a world action model, and PhysicalRSI \cite{mmlab2026physicalrsi} a VLA.

\FloatBarrier
\section{Results on each benchmark}
\label{app:results}

\subsection{RoboDojo}
\label{app:robodojo-results}

Five tasks stay below 17 of 30 with skills: store laptop and headphones (1), hang mugs (7), deposit coin (13), and fill pen holder and organize table (16 each). Among the Open tasks, store tools in toolbox passes no episode (\cref{app:toolbox}). Against Q0-Preview, the best leaderboard entry, the agent without skills is ahead on Generalization, Memory and Open and behind on Precision (52.5 against 58.5) and Long-Horizon (54.6 against 58.5). Against HexaShell-TaskTools it has a higher success rate on 36 of the 42 tasks, the same on four and a lower one on two, plug in charger (30\% against 32\%) and make kong (0 against 2\%); with skills it is higher on 40 tasks and the same on cover blocks (100\%) and store tools in toolbox (0).

\subsection{RoboCasa365}
\label{sec:robocasa}

\Cref{tab:robocasa} gives the success rates on the 50 target tasks: 18 atomic, 16 composite seen and 16 composite unseen. The overall rate is the mean over the tasks, which weights the three groups 18/16/16 as the leaderboard does. The leaderboard entries \cite{robocasa365leaderboard}, as of 1 November 2026, are policies trained on the pretraining data and evaluated on 50 episodes per task in the pretraining kitchens; the results from papers each use their own episodes; ours use the 1,500 test episodes of the pretrain50 split, 30 per task.

With nothing written for any task, HexaShell succeeds in 65.6\% of the atomic, 50.6\% of the composite seen and 61.0\% of the composite unseen test episodes, 59.3\% overall. The best leaderboard entry, Awomo-0.5I, reaches 64.4\% overall. The agent without skills is behind it on the atomic tasks (65.6 against 86.5) and the composite seen ones (50.6 against 58.8), and ahead on the composite unseen ones (61.0 against 45.2). Of its 610 failed episodes, the agent ended 393 itself, 215 ran out of steps and 2 reached the time limit.

With skills, HexaShell succeeds in 91.1\% of the atomic, 85.0\% of the composite seen and 80.0\% of the composite unseen test episodes, 85.6\% overall. On the same 1,500 episodes, the skills gain 466 and lose 72 ($p<0.001$). The largest gains are on tasks at the sink and on appliance doors and buttons: ScrubCuttingBoard (2 to 30 of 30), WashLettuce (1 to 26), LoadDishwasher (3 to 27), PreSoakPan (6 to 30), RinseSinkBasin (4 to 26), TurnOnSinkFaucet (7 to 30), CloseToasterOvenDoor (6 to 28) and TurnOnMicrowave (8 to 29). Six tasks pass one or two episodes more without skills. The run without skills kept up to 50 pictures in the conversation instead of 20, so the gap is not the skills' alone, though on RoboDojo a larger window made no measurable difference (\cref{sec:ablation}). The weakest tasks with skills are HeatKebabSandwich (10), PackIdenticalLunches (13), ArrangeTea (16), SteamInMicrowave (17), SeparateFreezerRack (18) and MakeIceLemonade (19). Of the 216 failed episodes with skills, the agent ended 159 itself and 57 ran out of steps.

With skills, raising the effort from high to xhigh lifts the successes from 1,229 to 1,284 of 1,500 (136 gained, 81 lost, $p<0.001$). Of the 50 tasks, 27 go up and 13 down. The largest gains are OpenCabinet (17 to 27 of 30), GarnishPancake (22 to 27), PackIdenticalLunches (8 to 13) and StoreLeftoversInBowl (20 to 25).

Every task has a skill written on learning seeds, so with skills the composite unseen tasks are not unseen to our agent, though their kitchens and seeds are. The result with skills is listed on the agentic track of the RoboCasa365 leaderboard (submitted on 31 October and added on 1 November 2026) as our own evaluation on pretrain50 seeds 20--49, not under the official protocol.

\subsection{LIBERO-PRO and RoboTwin~2.0}
\label{sec:libero-robotwin}

\begin{table}[!htbp]
\centering
\caption{\textbf{LIBERO-PRO results.} Top: successes by perturbation and suite, out of 100 episodes unless another count is given; with skills on the test initial states, without skills on other initial states. Bottom: success rate (\%) on the cells that other agent papers report, the mean of the per-cell rates, with our episodes restricted to the same cells; each paper uses its own episodes. $^*$Our mean of the paper's per-cell results.}
\label{tab:libero}
\small
\begin{tabular}{@{}llccccc@{}}
\toprule
 & Perturbation & Spatial & Object & Goal & LIBERO-10 & All \\
\midrule
\multirow{5}{*}{With skills} & Position & 97 & 96 & 99 & 94 & 386/400 \\
 & Object & 90 & 100 & 100 & 79/90 & 369/390 \\
 & Language & 96 & 100 & 98 & 96 & 390/400 \\
 & Task & 77/80 & 100 & 100 & 87/90 & 364/370 \\
 & All & 360/380 & 396/400 & 397/400 & 356/380 & \textbf{1,509/1,560 (96.7\%)} \\
\midrule
\multirow{5}{*}{Without skills} & Position & 87 & 92 & 76 & 70 & 325/400 \\
 & Object & 83 & 70 & 78 & 56/90 & 287/390 \\
 & Language & 82 & 82 & 84 & 80 & 328/400 \\
 & Task & 74/80 & 84 & 68 & 56/90 & 282/370 \\
 & All & 326/380 & 328/400 & 306/400 & 262/380 & 1,222/1,560 (78.3\%) \\
\bottomrule
\end{tabular}

\vspace{0.8em}
\begin{tabular}{@{}llc@{}}
\toprule
Cells & Method & Success (\%) \\
\midrule
\multirow{6}{*}{\shortstack[l]{Six: position and task\\perturbations of Spatial,\\Object and Goal}}
 & $\pi_{0.5}$, in \cite{fu2026capx} & 12.8$^*$ \\
 & RATs \cite{zhang2026rats} & 43.8 \\
 & RHO \cite{elmaaroufi2026rho} & 45.0 \\
 & Robo-Harness K1 with Gemini~3.7 Flash \cite{li2026roboharnessk1} & 77.2 \\
 & HexaShell without skills & 83.3 \\
 & HexaShell with skills & \textbf{98.0} \\
\midrule
\multirow{7}{*}{\shortstack[l]{Eight: the six, and\\the same two perturbations\\of LIBERO-10}}
 & $\pi_{0.5}$, in \cite{zhang2026harnessvla} & 11.0 \\
 & OmniBrain-0 RSI \cite{xiaomi2026omnibrain0} & 69.3 \\
 & Harness VLA, Codex \cite{zhang2026harnessvla} & 72.1 \\
 & GPT-5.6-Sol, as reported in \cite{xiaomi2026omnibrain0} & 78.3 \\
 & Harness VLA, Claude Code \cite{zhang2026harnessvla} & 82.4 \\
 & HexaShell without skills & 79.0 \\
 & HexaShell with skills & \textbf{97.4} \\
\midrule
40 tasks & PyRUA-Lean \cite{si2026pyrua} & 94.0 \\
\bottomrule
\end{tabular}
\end{table}

\begin{table}[!htbp]
\centering
\caption{\textbf{RoboTwin~2.0 results.} Success rate (\%) on the 50 tasks, in clean scenes and in randomized ones (``hard'' on the leaderboard). We ran only the randomized scenes: without skills in the learning round, and with skills on the test, with other seeds and unseen instructions.}
\label{tab:robotwin}
\small
\begin{tabular}{@{}lcc@{}}
\toprule
Method & Clean & Randomized \\
\midrule
\multicolumn{3}{@{}l}{\emph{Leaderboard}} \\
GO-3 Nano Preview \cite{agibot2026go3nano} & 88.5 & 81.5 \\
PrismWAM (uses depth) & 91.8 & 75.5 \\
$\pi_{0.5}$ \cite{intelligence2025pi05} & 70.7 & 46.0 \\
\midrule
\multicolumn{3}{@{}l}{\emph{Reported in papers}} \\
GPT-5.6-Sol, few-shot with VLA skills, as reported in \cite{xiaomi2026omnibrain0} & 92.0 & 90.4 \\
OmniBrain-0 RSI \cite{xiaomi2026omnibrain0} & 90.4 & 88.4 \\
RoboDawn, GPT-6-Astra, one-shot / zero-shot \cite{guo2026robodawn} & -- & 73.6 / 53.2 \\
Harness VLA, Claude Code / Codex \cite{zhang2026harnessvla} & -- & 58.4 / 58.0 \\
Robo-Harness K1 \cite{li2026roboharnessk1} & 32.0 & 28.0 \\
\midrule
HexaShell, no skills (learning round) & -- & 82.2 \\
\textbf{HexaShell, with skills (test)} & -- & \textbf{96.8} \\
\bottomrule
\end{tabular}
\end{table}

On LIBERO-PRO, with skills, HexaShell succeeds in 1,509 of the 1,560 test episodes (96.7\%, \cref{tab:libero}). The 71 tasks with skills pass 696 of 710 and the 85 without pass 813 of 850. Without any skills, on other initial states, the agent passes 78.3\%. The test episodes use initial states 20--29. The run without skills uses initial states 10--19: for the 102 tasks without a skill at that point it is the validation round, and for the other 54 a control run with the skills switched off. The difference is largest when the perturbation changes the objects or the task, and on the Goal and LIBERO-10 suites. The paired control of \cref{tab:ablation} isolates the skills on 54 tasks: 252 of 540 episodes succeed without them and 531 with them. Most papers report a few of LIBERO-PRO's cells. On the six cells used by CaP-X \cite{fu2026capx}, HexaShell reaches 98.0\% with skills and 83.3\% without, against 77.2\% for the best other agent we found. On the eight cells of Harness VLA \cite{zhang2026harnessvla} it reaches 97.4\% and 79.0\%, against 82.4\%. There, without skills, it is below Harness VLA with Claude Code. PyRUA-Lean reports 94.0\% on 40 tasks, with no breakdown by cell \cite{si2026pyrua}.

Of the 51 failed episodes, 50 ran out of steps or ended for lack of them, 38 at the limit and 12 with \texttt{finish(no\_budget)}; the other ended with \texttt{stuck}. One task, LIBERO-10's task~3 under the object perturbation, failed all 10 episodes in all three rounds: there the perturbed bowl is too wide for the drawer to close.

On RoboTwin~2.0, in randomized scenes with unseen instructions, HexaShell succeeds in 726 of the 750 test episodes (96.8\%, \cref{tab:robotwin}), and 36 of the 50 tasks pass all 15. The leaderboard entries of \cref{tab:robotwin}, as of 30 October 2026, are policies trained on clean demonstrations and evaluated on 100 episodes per task in each setting; the results from papers are agents, each with its own episodes. In the randomized setting the top leaderboard entry, GO-3 Nano Preview, reaches 81.5\%. The best result in a paper is GPT-5.6-Sol with skills for a VLA, at 90.4\% as reported by OmniBrain-0. Without skills, the learning round already passed 411 of 500 episodes (82.2\%). Its seeds and instructions differ from the test, so it is not a paired control, and we did not run one on the test seeds. The skills were revised once, for the 15 tasks that failed in the validation round. On the same validation seeds these tasks went from 126 to 142 of 150 (in-sample). On the test they pass 205 of 225, and the other 35 tasks pass 521 of 525. The weakest tasks are stacking three bowls (11 of 15), and hanging a mug, putting bottles in the dustbin and scanning an object (12 each). Of the 24 failures, the agent ended 14 itself, 7 reached the reply limit and 3 ran out of steps.

\begin{figure}[H]
\centering
\includegraphics[width=\linewidth]{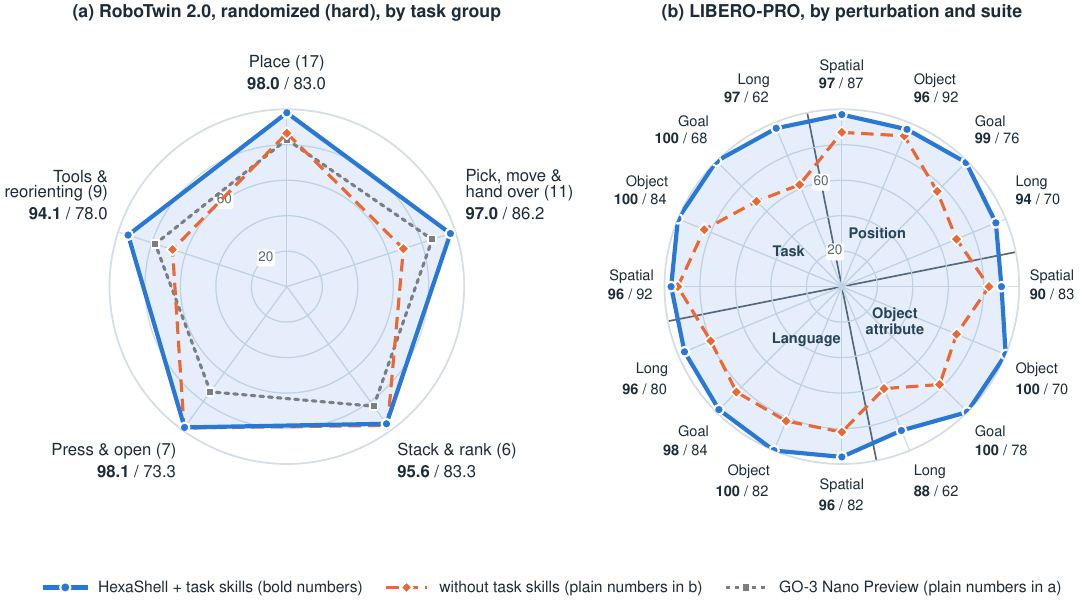}
\caption{\textbf{LIBERO-PRO and RoboTwin~2.0 in detail}, success in \%. Solid line and bold numbers: HexaShell with skills; dashed line: without. \textbf{(a)}~RoboTwin~2.0, randomized setting, tasks grouped by verb (number of tasks in brackets); dotted line and plain numbers: GO-3 Nano Preview \cite{agibot2026go3nano} on the leaderboard \cite{robotwin2leaderboard}. \textbf{(b)}~LIBERO-PRO by perturbation (inside) and suite (Long is LIBERO-10); plain numbers are without skills. Groups and episodes are given in \cref{sec:libero-robotwin}.}
\label{fig:radar-detail}
\end{figure}

\Cref{fig:radar-detail} breaks both benchmarks down, by task group on RoboTwin~2.0 and by perturbation and suite on LIBERO-PRO. The RoboTwin~2.0 groups follow the verb in the task name. Place: the \texttt{place\_*} tasks. Pick, move \& hand over: \texttt{pick\_*}, \texttt{move\_*}, \texttt{handover\_*}, \texttt{put\_bottles\_dustbin}, \texttt{grab\_roller} and \texttt{lift\_pot}. Stack \& rank: \texttt{stack\_*} and \texttt{blocks\_ranking\_*}. Press \& open: \texttt{click\_*}, \texttt{press\_stapler}, \texttt{turn\_switch}, \texttt{open\_*} and \texttt{put\_object\_cabinet}. Tools \& reorienting: the other nine. The line without skills is the learning round, 10 episodes per task on other seeds with seen instructions. GO-3 Nano Preview is the mean of its per-task clean2random results on the leaderboard, 100 episodes per task.

\FloatBarrier
\section{Per-task results}
\label{app:pertask}

\Cref{tab:pertask-robodojo,tab:pertask-robocasa,tab:pertask-robotwin} give the successes of every task. LIBERO-PRO has 156 tasks of 10 episodes each; \cref{tab:libero} and \cref{fig:radar-detail} break it down by suite and perturbation.

\Cref{tab:pertask-robodojo} counts the successes of each RoboDojo task in its 30 test episodes (seeds 0--2, layouts 10--19), in the two runs of \cref{tab:main}; in the twelve Generalization tasks, half of the episodes add clutter. The Open tasks have no skills, so both columns share one run of them. \Cref{tab:pertask-robotwin} lists the 14 RoboTwin~2.0 tasks that failed at least one of their 15 test episodes, in randomized scenes with unseen instructions; the other 36 tasks passed all 15.

\begin{table}[!htbp]
\centering
\caption{\textbf{RoboDojo per task.} Successes in 30 test episodes, without and with skills. $^*$Not run again after passing none of 30 in an earlier run without skills; counted as 0.}
\label{tab:pertask-robodojo}
\small
\begin{minipage}[t]{0.48\linewidth}
\vspace{0pt}\centering
\begin{tabular}{@{}lrr@{}}
\toprule
Task & No skills & Skills \\
\midrule
\multicolumn{3}{@{}l}{\emph{Generalization}} \\
arrange largest number & 26 & 26 \\
fold clothes & 22 & 23 \\
hang mugs & 2 & 7 \\
make toast & 15 & 21 \\
pack objects into box & 8 & 26 \\
pour liquid into cup & 17 & 29 \\
push T & 30 & 30 \\
sort nesting dolls by size & 25 & 29 \\
stack blocks & 30 & 30 \\
stack bowls & 30 & 30 \\
store laptop and headphones & 0 & 1 \\
sweep blocks & 27 & 28 \\
\midrule
\multicolumn{3}{@{}l}{\emph{Precision}} \\
build tower & 14 & 29 \\
deposit coin & 13 & 13 \\
fasten screws & 10 & 30 \\
insert key & 0 & 20 \\
insert tubes & 24 & 29 \\
play xylophone & 28 & 30 \\
plug in charger & 9 & 28 \\
pour balls into vase & 28 & 28 \\
\bottomrule
\end{tabular}
\end{minipage}\hfill
\begin{minipage}[t]{0.48\linewidth}
\vspace{0pt}\centering
\begin{tabular}{@{}lrr@{}}
\toprule
Task & No skills & Skills \\
\midrule
\multicolumn{3}{@{}l}{\emph{Long-Horizon}} \\
classify objects & 30 & 29 \\
fill egg holder & 4 & 19 \\
fill pen holder & 2 & 16 \\
make kong & 0 & 30 \\
organize table & 10 & 16 \\
play stacking toy & 26 & 30 \\
play tic-tac-toe & 29 & 30 \\
put bottles into dustbin & 30 & 29 \\
\midrule
\multicolumn{3}{@{}l}{\emph{Memory}} \\
cover blocks & 30 & 30 \\
imitate sorting sequence & 29 & 30 \\
match and pick from conveyor & 30 & 30 \\
press by number & 29 & 30 \\
swap T & 30 & 30 \\
swap blocks & 30 & 28 \\
\midrule
\multicolumn{3}{@{}l}{\emph{Open (no skills)}} \\
align blocks & 29 & 29 \\
classify objects by language & 28 & 28 \\
general pickup & 29 & 29 \\
pick from conveyor by image & 20 & 20 \\
pour by language & 28 & 28 \\
solve equation & 27 & 27 \\
stack blocks by language & 29 & 29 \\
store tools in toolbox & 0$^*$ & 0$^*$ \\
\bottomrule
\end{tabular}
\end{minipage}
\par\smallskip
All 42 tasks: 857/1,260 without skills; 1,054/1,260 with skills.

\end{table}

\begin{table}[!htbp]
\centering
\caption{\textbf{RoboCasa365 per task.} Successes out of 30 test episodes, without and with skills.}
\label{tab:pertask-robocasa}
\small
\begin{tabular}{@{}lrr@{\hspace{2.5em}}lrr@{}}
\toprule
Task & No skills & Skills & Task & No skills & Skills \\
\midrule
\multicolumn{3}{@{}l}{\emph{Atomic}} & \multicolumn{3}{l}{\emph{Composite seen}} \\
CloseBlenderLid & 15 & 21 & DeliverStraw & 23 & 28 \\
CloseFridge & 20 & 26 & GetToastedBread & 25 & 25 \\
CloseToasterOvenDoor & 6 & 28 & KettleBoiling & 29 & 28 \\
CoffeeSetupMug & 15 & 26 & LoadDishwasher & 3 & 27 \\
NavigateKitchen & 25 & 26 & PackIdenticalLunches & 9 & 13 \\
OpenCabinet & 14 & 27 & PreSoakPan & 6 & 30 \\
OpenDrawer & 22 & 29 & PrepareCoffee & 21 & 24 \\
OpenStandMixerHead & 14 & 28 & RinseSinkBasin & 4 & 26 \\
PickPlaceCounterToCabinet & 23 & 26 & ScrubCuttingBoard & 2 & 30 \\
PickPlaceCounterToStove & 29 & 28 & SearingMeat & 20 & 28 \\
PickPlaceDrawerToCounter & 24 & 28 & SetUpCuttingStation & 21 & 25 \\
PickPlaceSinkToCounter & 28 & 27 & StackBowlsCabinet & 29 & 28 \\
PickPlaceToasterToCounter & 25 & 27 & SteamInMicrowave & 10 & 17 \\
SlideDishwasherRack & 24 & 28 & StirVegetables & 20 & 28 \\
TurnOffStove & 27 & 28 & StoreLeftoversInBowl & 20 & 25 \\
TurnOnElectricKettle & 28 & 30 & WashLettuce & 1 & 26 \\
\cmidrule(r){4-6}
TurnOnMicrowave & 8 & 29 & Composite seen, of 480 & 243 & 408 \\
TurnOnSinkFaucet & 7 & 30 & & & \\
\cmidrule(r){1-3}
Atomic, of 540 & 354 & 492 & & & \\
\midrule
\multicolumn{3}{@{}l}{\emph{Composite unseen}} & & & \\
ArrangeBreadBasket & 25 & 23 & MakeIceLemonade & 19 & 19 \\
ArrangeTea & 18 & 16 & PanTransfer & 24 & 25 \\
BreadSelection & 23 & 28 & PortionHotDogs & 27 & 27 \\
CategorizeCondiments & 24 & 26 & RecycleBottlesByType & 30 & 30 \\
CuttingToolSelection & 15 & 30 & SeparateFreezerRack & 3 & 18 \\
GarnishPancake & 14 & 27 & WaffleReheat & 22 & 25 \\
GatherTableware & 10 & 23 & WashFruitColander & 9 & 30 \\
HeatKebabSandwich & 6 & 10 & WeighIngredients & 24 & 27 \\
\cmidrule(r){1-3}
Composite unseen, of 480 & 293 & 384 & & & \\
\midrule
All 50 tasks, of 1,500 & 890 & 1,284 & & & \\
\bottomrule
\end{tabular}
\end{table}

\begin{table}[!htbp]
\centering
\caption{\textbf{RoboTwin~2.0 per task.} Successes in 15 test episodes of the 14 tasks that failed at least one; the other 36 tasks pass all 15. $^*$Skill revised after validation.}
\label{tab:pertask-robotwin}
\small
\begin{tabular}{@{}lr@{\hspace{2.5em}}lr@{}}
\toprule
Task & /15 & Task & /15 \\
\midrule
stack\_bowls\_three$^*$ & 11 & pick\_dual\_bottles$^*$ & 14 \\
hanging\_mug$^*$ & 12 & place\_bread\_skillet$^*$ & 14 \\
put\_bottles\_dustbin$^*$ & 12 & place\_dual\_shoes$^*$ & 14 \\
scan\_object$^*$ & 12 & place\_bread\_basket & 14 \\
put\_object\_cabinet$^*$ & 13 & place\_object\_basket & 14 \\
dump\_bin\_bigbin$^*$ & 14 & place\_phone\_stand & 14 \\
pick\_diverse\_bottles$^*$ & 14 & rotate\_qrcode & 14 \\
\bottomrule
\end{tabular}
\end{table}

\subsection{Six tasks with and without skills}
\label{app:cases}

The tables above count successes; the rest of this section shows how a skill changes an episode. Each case follows one RoboDojo test episode that fails in the run without skills of \cref{tab:pertask-robodojo} and succeeds in the run with them. The two runs share the model, the reasoning effort, the code and the image window, so the skill is the only difference.

Of the 1,020 episodes outside Open, 234 succeed only with skills, in 25 tasks, and 37 only without, in 16 (\cref{sec:knowledge}). We took six of the tasks with the most such gains, two from each of the Generalization, Precision and Long-Horizon dimensions, leaving out plug in charger, which \cref{sec:cases} describes. Together they account for 110 of the gains and 7 of the losses; the Memory tasks gained 2 episodes in all. For each task we sorted the failures without the skill by their cause, chose an episode with a frequent cause that passed with the skill, and read both of its records in full: the code and output of every cell and the camera images after it. Positions and heights are those that the agent measured or that the arms reported. The records hold no poses of objects, so where an object ended is judged from the images. \Cref{tab:cases-dojo} summarizes the six cases.

\begin{table}[!htb]\centering\small\setlength{\tabcolsep}{4pt}\renewcommand{\arraystretch}{1.1}
\caption{\textbf{Six RoboDojo test episodes, without and with the skill.} Each row is one episode: what went wrong without the skill, what the skill supplies, the step at which the episode succeeded with the skill, out of its budget, and the task's successes in 30 test episodes without and with skills.}
\label{tab:cases-dojo}
\begin{tabularx}{\textwidth}{@{}>{\raggedright\arraybackslash}p{2.2cm} c c >{\raggedright\arraybackslash}X >{\raggedright\arraybackslash}X r c@{}}\toprule
Task & Seed & Layout & Without the skill & The skill supplies & Step & Successes \\ \midrule
\multicolumn{7}{@{}l}{\emph{Generalization}} \\
Make toast & 0 & 11 & released the second slice leaning in its slot; it fell flat while being straightened, and the steps ran out & one arm and one pitch, checked by dry runs, kept from rack to slot & 222/1,400 & 15 $\to$ 21 \\ \addlinespace[2pt]
Pack objects into box & 0 & 14 & gripped the box to straighten it; the box tipped, and the steps ran out & when a moved box still counts; leave it alone; centre from the rim corners & 614/1,300 & 8 $\to$ 26 \\ \midrule
\multicolumn{7}{@{}l}{\emph{Precision}} \\
Build tower & 2 & 10 & one arm set both lower pillars, both from the left half; the finished tower scored 10 & one pillar from each half of the table in each pillar level & 545/1,050 & 14 $\to$ 29 \\ \addlinespace[2pt]
Fasten screws & 0 & 12 & turned the first nut both ways; still on the bolt's tip after eleven turns, and the steps ran out & a clockwise turn that pushes down to 0.795\,m; seated and tip heights & 441/1,900 & 10 $\to$ 30 \\ \midrule
\multicolumn{7}{@{}l}{\emph{Long-Horizon}} \\
Fill pen holder & 2 & 15 & turned two clip pens writing tip up; four pens in and arms home, at a score of 25 & which end of each pen goes up, and the roll that raises it & 482/1,100 & 2 $\to$ 16 \\ \addlinespace[2pt]
Make kong & 0 & 11 & moved the opponent's discard six times and never drew a tile from the wall & the tile to place comes from the wall; where it stands; a hand-over & 375/600 & 0 $\to$ 30 \\
\bottomrule\end{tabularx}
\end{table}

\newcommand{\casefig}[5]{%
\begin{tikzpicture}[
  x=1mm, y=1mm, font=\sffamily\fontsize{6.2}{7.4}\selectfont, text=rsink,
  hd/.style={font=\sffamily\bfseries\fontsize{7}{8.4}\selectfont, anchor=base west, inner sep=0pt},
  img/.style={anchor=north west, inner sep=0pt},
  txt/.style={anchor=north west, inner sep=0pt, align=left, text width=38.6mm},
]
\node[hd] at (0,2.2) {Without the skill};
\node[hd] at (84.7,2.2) {With the skill};
\draw[rsgray, line width=0.4pt] (0,1.2) -- (79.2,1.2) (84.7,1.2) -- (163.9,1.2);
\foreach \x/\f/\t in {0/a/{#2}, 40.6/b/{#3}, 84.7/c/{#4}, 125.3/d/{#5}} {
  \node[img] (p) at (\x,0) {\includegraphics[width=38.6mm]{fig/img/case_#1_\f.jpg}};
  \node[txt] at ($(p.south west)+(0,-1.2)$) {\t};
}
\end{tikzpicture}}

\subsection{Make toast: one pitch from rack to slot}
\label{app:case-toast}

In make toast the robot puts two of four bread slices from a rack into the two slots of a toaster and presses the lever down; full credit also requires exactly two slices left in the rack and the robot back at its origin. On test seed~0, layout~11, the agent without the skill asked the right arm to lift each slice at pitch $60^\circ$, a pose just out of its reach, and the arm went to the closest reachable pose, which, the tool noted, ``tilts what it holds by $5^\circ$'' (cells 8 and 14). The second slice shifted in the fingers on its way down to the left slot; the descent stopped 13\,mm short, where the hand opened and left the slice leaning (cells 15--16). While being straightened, it slipped out and fell flat beside the toaster (cell 39). A third slice from the rack took its slot (cells 60--63). The eleventh close on the fallen slice lifted it, at step 1,168, and the agent set it back in the rack. The episode ended at step 1,393 of 1,400, with the slice in the left slot tilted and a partial score of 25 (\cref{fig:case-toast}).

\begin{figure}[!htb]
\centering
\casefig{toast}{\textbf{(a)} Cells 16--17: released where its descent stopped 13\,mm short, the second slice leans across the left slot.}{\textbf{(b)} Cells 39--40: the slice slips out while being straightened and falls flat beside the toaster, with 936 steps left.}{\textbf{(c)} Cell 19: both slices, held at pitch $45^\circ$ from rack to slot, stand edge-on in their slots.}{\textbf{(d)} Cell 21: a closed hand presses the lever at pitch $30^\circ$; success at step 222, during \texttt{robot.home()}.}
\caption{\textbf{Make toast, test seed 0, layout 11: a leaning slice and two seated ones.} Right wrist camera in (a), above the toaster, and in (c), above its lever; head camera in (b) and (d); all cropped. In both runs the first slice went into the right slot.}
\label{fig:case-toast}
\end{figure}

The skill, 65 lines without code, settles the carry before the first grasp. Dry runs are to find ``ONE arm and pitch'' that reach both a rack slice, with its lift of about 11\,cm, and the slots, trying pitch $60^\circ$ first, as in all three development successes, then $45^\circ$ and $30^\circ$. It warns that ``an offered pitch or yaw tilts or turns the held slice'', and that slices kept at the same pitch and roll from rack to slot stayed upright, ``6/6 seated''. It also records that a lost slice is hard to make up for: ``A flat slice was seated once in 44 low closes (6 runs)'', and a third slice from the rack, tried in four development runs, scored 0 in each.

With the skill, on the same layout, dry runs in cell 7 found that at pitch $60^\circ$ the lift could only be approached, $7^\circ$ off, while at $45^\circ$ the grasp, the lift and the slot were all reachable. Both slices then stayed within $1^\circ$ of pitch $45^\circ$ and roll $0^\circ$ from rack to slot and went in without a block (cells 8--19); success came during \texttt{robot.home()} at step 222 of 1,400. Over the 30 test episodes, successes rise from 15 to 21. In 14 of the 15 failures without the skill a slice fell, in 8 of them, as here, while the agent was correcting a slice that rested tilted, high or on the rim. With the skill, a slice slipped in each of the 9 failures as well. Three of the 4 episodes that passed only without the skill have cluttered layouts. There the agent with the skill judged the rack and the toaster too far apart for a steep carry with one arm, a case in which four development runs had all failed, and lost a slice carried at a pitch of $0^\circ$ to $30^\circ$. Without the skill, the agent handed the slices between the arms, which the skill advises against: hand-overs passed none of 14 slices in the development runs. In the fourth, the second slice tipped the first during the carry.
\subsection{Pack objects into box: a box that needed no straightening}
\label{app:case-pack}

In pack objects into box the robot puts a toy car, an electric toothbrush, a hammer and a shoe into an open box, each with its front to the left; full credit requires all four in ``the aligned box'' and a return to origin. On test seed~0, layout~14, the box stood about $20^\circ$ off the table's edges, and the agent without the skill wrote ``I'll straighten it before packing''. Its left hand gripped the near wall, lifted the box and turned it (cells 7--9), then again (cells 14--15). After the arm carrying the car shifted the box, the agent dragged it by the rim to the middle of the table (cells 23--24). There, picking up the shoe beside it tipped the box onto its end (cell 35); righting it took 242 steps (cells 38--50). The cells that handled the box took 500 of the 1,300 steps; the budget ran out at a score of 10, one item placed correctly (\cref{fig:case-pack}).

\begin{figure}[!htb]
\centering
\casefig{pack}{\textbf{(a)} Cell 8: to straighten the box, the left hand lifts it by its near wall.}{\textbf{(b)} Cell 35: picking up the shoe beside it tips the box on end; righting it takes 242 steps.}{\textbf{(c)} Cell 15: hammer, car and toothbrush have gone straight down from 0.95\,m into the untouched box.}{\textbf{(d)} Cell 20: the shoe goes in last, and success comes during \texttt{robot.home()}, at step 614 of 1,300.}
\caption{\textbf{Pack objects into box, test seed 0, layout 14: a box gripped and a box left alone.} Head camera, cropped. In (c) and (d) the box stands where it started, about $20^\circ$ off the table's edges.}
\label{fig:case-pack}
\end{figure}

The skill, 63 lines without code, states what ``aligned'' meant in the development runs: ``A tipped or far-turned box counts nothing (the card: aligned); shifted but upright, items counted (4 successes).'' For a moved box it records ``Gripping, hooking or dragging a moved box: 13 runs, none restored'' and prescribes ``Upright, roughly left-right, reachable: new centre, go on''. It has the centre taken once, from the rim corners in a head image ``with both arms clear of the box'', and each item carried at a grasp height of 0.93--0.95\,m over the near wall, then straight down: the shoe crossing the wall at 0.89--0.90\,m ``pushed the box 20--40\,cm (2 runs)''.

With the skill, on the same layout, no hand touched the box. Warned in cell 3 that three of the pixels it had picked lay ``on the robot's own arm or gripper'', the agent moved the left arm aside and took the four rim corners (cells 7--8). The left arm placed all four items, two of them laid down for it by the right arm. Each item crossed the near wall at 0.95--0.96\,m and went straight down, the shoe last, and success came during \texttt{robot.home()} at step 614 of 1,300. Over the 30 test episodes, the task passed 8 times without the skill and 26 with it. Without the skill the agent gripped the box in all 22 failures and in 6 of the 8 successes; in 17 failures it did so before grasping any item, and 21 failures ended for lack of steps. The agent rarely mistook a front: in the 10 failures in which it said where the items' fronts were, it agreed with the skill in all but one. With the skill the box was gripped in 5 cluttered episodes, never before the first item; the 4 failures, also cluttered, each left the shoe outside, facing the wrong way or partly upright.
\subsection{Build tower: a lower pair taken from one side of the table}
\label{app:case-tower}

In build tower the robot stacks two pillars, a long board, two more pillars, a short board, a flat plate and a green prism. The task card gives a score of 10 for the lower layer, 30 with the middle board, and 100 for the full tower, each layer centred and both arms back at their start poses. On test seed~2, layout~10, without the skill, the right arm held the long board aside while the left arm set both lower pillars where the board had lain, at $x=-13$ and $+13$\,cm (cells 9 and 11). Both had started left of the table's middle. By cell 31 all eight pieces were stacked and both arms were home, at step 635 of 1,050, but the episode did not end. The agent turned the plate back to its starting orientation and then took the top apart to ``try a three-level tower''. It stopped at step 1,029 with a score of 10 (\cref{fig:case-tower}).

\begin{figure}[!htb]
\centering
\casefig{tower}{\textbf{(a)} Cell 13: the long board rests on both left-hand pillars; the right-hand pair waits for the upper level.}{\textbf{(b)} Cell 31: all eight pieces stacked and both arms home at step 635 of 1,050; no success follows.}{\textbf{(c)} Cell 11: the long board rests on one pillar from each side; one more waits on each side.}{\textbf{(d)} Cell 31: the finished tower; success comes 12 steps into \texttt{home()}, at step 545 of 1,050.}
\caption{\textbf{Build tower, test seed 2, layout 10: where the lower pillars came from.} Head camera, cropped. Left-hand and right-hand pillars started left and right of the table's middle; the finished towers in (b) and (d) look alike.}
\label{fig:case-tower}
\end{figure}

The card does not say which pillar goes where. The skill does: ``Build each pillar level from one pillar of each side: one that started left of the table's middle, one right.'' It cites three development runs in which one arm placed both lower pillars; none got past the lowest score, although two of them had finished the tower. Its list of common failures adds: ``Finished tower, no credit: a one-side lower pair. No cheap repair; prevent it.'' Its flow puts the tower centre 7--8\,cm farther from the robot than the long board's start, has each arm bring one lower pillar from its own side, and keeps the board's landing spot 5\,cm clear of the spare pillars: in the development runs, a pillar fetched within 5\,cm of the board hit it in 5 of 9 fetches, and one fetched from farther away in none of 18.

With the skill, on the same layout, the agent announced before its first motion that it would use ``one pillar from each side in both support levels''. It set the tower centre about 8\,cm farther from the robot than the long board's start, and with the board still on the table the left and right arms set their lower pillars at $x=-12.4$ and $+12.6$\,cm (cells 5 and 7). The left arm then laid the board across them. Success came 12 steps into the final \texttt{home()}, at step 545 of 1,050, after 31 cells instead of 45. Over the 30 test episodes the task passed 14 times without the skill and 29 times with it. Five of the 16 failures without the skill had taken both lower pillars from one side; all five stacked all eight pieces and scored 10. No success without the skill and no episode with it had such a pair. A one-side pair was not the only reason a finished tower went uncredited: in five more failures the agent stacked all eight pieces with one pillar from each side and no success followed, once with a lower pillar, in the agent's words, about 3\,cm off centre; four of these then lost a top piece while adjusting the tower. Five of the remaining six reported disturbing the lower board while fetching a pillar beside it, which the 5\,cm clearance addresses. The one failure with the skill also finished the tower with one pillar from each side, and stopped at a score of 30 for a reason the record does not show.
\subsection{Fasten screws: a clockwise turn that pushes down}
\label{app:case-screws}

In fasten screws the robot screws each of three nuts onto the bolt of its colour, fixed upright on the table (the task card calls the bolts screws). A nut counts once rotated onto its bolt to the required depth with the gripper open; success needs all three and both arms at home. Without the skill, on test seed~0, layout~12, the agent first turned the yellow nut $90^\circ$ clockwise, seen from above, to a grasp height of 0.804\,m; its top then measured 0.833\,m. The next four turns went counter-clockwise, and the agent wrote: ``The nut is rotating, but it hasn't visibly moved farther down the screw.'' At step 1,398, after eleven turns, the top still read 0.830\,m, and the agent suspected that ``the screw may be turning with the nut''. In all it turned this nut 16 times, $1{,}590^\circ$ clockwise and $1{,}760^\circ$ counter-clockwise, with no clockwise turn ending below 0.804\,m. It stopped at step 1,880 of 1,900 with two pairs fastened, a score of 50 (\cref{fig:case-screws}).

\begin{figure}[!htb]
\centering
\casefig{screws}{\textbf{(a)} Cell 9: the first turn, $90^\circ$ clockwise, ends at 0.804\,m; the nut's top then measures 0.833\,m.}{\textbf{(b)} Cell 39: after eleven turns, six counter-clockwise, the left hand, sent to hold the bolt still, is blocked.}{\textbf{(c)} Cell 6: one $120^\circ$ clockwise turn, pushing down to 0.795\,m; the gripper then opens.}{\textbf{(d)} Cell 19: all three nuts seated, with measured tops of 0.824--0.827\,m; success at step 441 of 1,900.}
\caption{\textbf{Fasten screws, test seed 0, layout 12: on the tip or seated.} (a, c) Right wrist camera at the end of the first turn on the yellow nut, (a) before the gripper opens and (c) after; (b, d) head camera, cropped. Turn heights are heights of the grasp point.}
\label{fig:case-screws}
\end{figure}

The skill says the ``screws'' are ``bolts fixed to the table'' and that a nut ``resting on the bolt's tip'' does not count. Its step~4 makes the turn one \texttt{goto} with \texttt{yaw=-120} and \texttt{contact=True} to a grasp height of 0.795\,m, ``clockwise from above while pushing down''; step~5 repeats it after a regrip. In the development runs, such a pushed turn ``seated or read seated 8 of 11 first placements'', while ``first turns ending above 0.800 left 5 of 8 nuts on the tip''. ``Turning without pushing is slow'', about $1{,}400^\circ$ for one nut in one run, ``and counter-clockwise turns undo it''. The skill also gives the measured tops of a seated nut, 0.821--0.827\,m, and of one on the tip, 0.829--0.836\,m; both yellow-nut readings above fall in the second range.

With the skill, on the same layout, the agent made this turn with the skill's call in cell~6 (\cref{fig:case-screws}c). After step~5 repeated the turn, the nut's top measured 0.824\,m, and the agent wrote: ``The yellow nut is centred and reads at the seated height after two tightening turns.'' The red and blue nuts followed alike, reading 0.827 and 0.825\,m, and the episode succeeded at step 441 of 1,900. Over the 30 test episodes the task rose from 10 successes to 30. All 20 failures without the skill ran out of steps in the same way: each turned nuts both ways (150 of 474 turns counter-clockwise), 240 of the 324 clockwise turns ended above 0.800\,m, and the hands spent 551--1,535 steps at a single bolt. Even the ten successes needed 1,260--1,809 steps. With the skill, all 90 first turns were the pushed turn of step~4, all 385 turns were clockwise, and episodes took 419--1,511 steps.
\subsection{Fill pen holder: the end that goes up}
\label{app:case-pens}

In fill pen holder, one hand holds a pen holder while the other fills it with four pens. The task card gives 10 points for a pen ``correctly inserted tip-up into the upright pen holder'', 25 and 40 for two and three, and success needs all four in, the holder upright and both arms home. On test seed~2, layout~15, the pens are two highlighters and a black and a white pen, each with a clip. Without the skill, the agent picked the black and white pens at yaw $0^\circ$ and $-8^\circ$, turned them to roll $178$--$180^\circ$, writing tip up, and reported them ``in the holder tip-up''. At step 761 all four pens were in and both arms home, with 339 steps left, yet no success came. The agent then turned the purple highlighter over but never went back to the black or white pen, and stopped at step 1,068 of 1,100 with a best score of 25, the value for two pens (\cref{fig:case-pens}).

\begin{figure}[!htb]
\centering
\casefig{pens}{\textbf{(a)} Cell 8: the black pen, picked at yaw $0^\circ$ and turned with roll $180^\circ$, writing tip up.}{\textbf{(b)} Cell 55: four pens in, yet no success after both arms were home at step 761; black and white pens writing tip up.}{\textbf{(c)} Cell 13: the white pen, picked at yaw $-8^\circ$ and turned with roll $0^\circ$, clip end up.}{\textbf{(d)} Cell 22: holder set down, four pens clip end up; success at step 482, during \texttt{robot.home()}.}
\caption{\textbf{Fill pen holder, test seed 2, layout 15: writing tip up or clip end up.} Head camera in (a), left wrist camera in (b) and (c), right wrist camera in (d); (a), (b) and (d) cropped. Both runs pick the black and white pens at the same yaw, within $1^\circ$, and differ in the roll that turns them upright.}
\label{fig:case-pens}
\end{figure}

The skill replaces ``tip-up'' with a rule, ``A pen counts only CLIP END UP'': highlighters and the gold-clip pen go cap up, but ``click and gel pens writing tip DOWN'', citing all three winning development runs and ``tip up 0/8''. It states which roll raises the clip (``Roll 0 raises the end the yaw heads to'', ``roll 180 the other'') and asks for a left wrist image, which costs no steps: ``clip end up?''. Its final checks cover the state of step 761: ``Four in, arms home, holder down, no success: a pen is wrong end up or not down''.

With the skill, the agent gripped the holder at pitch $5^\circ$, the skill's fallback when $0^\circ$ is refused. It picked the black and white pens at nearly the same yaws but kept roll $0^\circ$, and each left wrist image showed the clip end up; every pen went in at the first attempt. As the skill prescribes, it set the holder down, moved the right hand with \texttt{by=(0, -0.08, 0.10)} and called \texttt{robot.home()} once: success at step 482 of 1,100, in 22 cells against 74. Over the 30 test episodes, successes rise from 2 to 16. In 20 of the 28 failures without the skill, the score stayed below the number of pens the agent reported in the holder, 13 times with all four in. The agent wrote of placing pens tip up in 25 of the 28, while the two successes had only gold-clip pens and highlighters, which count cap up. All 14 failures with the skill ran out of steps with a highlighter that had slipped, caught on the rim or fallen out of the held holder; the skill itself records that, inserted at the yaw it prescribes, ``highlighters stayed 8 of 16''.
\subsection{Make kong: a target tile that the task card does not identify}
\label{app:case-kong}

In make kong an opponent arm pushes out a discard, and the robot must ``declare a kong with the matching tiles''. The robot's row of standing tiles ends in a pair; the wall, two stacks of two tiles, lies to the left. The score table requires that the three matching tiles are pushed down, the others stay upright and ``the target tile is correctly grasped and placed at the required position'', without saying which tile or where. Without the skill, on test seed~0, layout~11, the agent pushed the three matching tiles face up by step 202. Planning to ``bring over the discard'', it picked the discard up in cell~12, then moved it between the middle tile and the table in front of the three: ``The discard is on the middle tile, but the declaration has not registered.'' After six set-downs, the 600-step budget ran out in cell~30, during a third homing. Neither arm reached for the wall (\cref{fig:case-kong}).

\begin{figure}[!htb]
\centering
\casefig{kong}{\textbf{(a)} Cell 12: with the three maple-leaf tiles face up, the left arm lifts the opponent's discard.}{\textbf{(b)} Cell 30: the budget runs out with the discard partly on the middle tile; the wall is untouched.}{\textbf{(c)} Cell 13: the left arm lifts the nearer wall stack's top tile; the discard lies untouched.}{\textbf{(d)} Cell 24: the right arm lowers the tile, on end, 5\,cm beyond the outer pair tile's centre.}
\caption{\textbf{Make kong, test seed 0, layout 11: the discard or a tile from the wall.} Head camera, cropped. The wall is the two stacks at the left of (b) and (c); opening the right gripper after (d) ends the episode in success.}
\label{fig:case-kong}
\end{figure}

The skill names the tile: ``The tile to place is the one you draw from the wall, not the discard: leave the discard where it lies (10 runs that moved it all scored 0).'' It also names the place: the top tile of the nearer wall stack ``stands on end one tile pitch past the outer pair tile, on the row line''. Since only the left arm reaches the wall and only the right arm the target, the skill's steps 4--9 hand the tile over: the left arm sets it down, grips it again across its width and stands it on end, and the right arm lowers it 5\,cm beyond the outer pair tile's measured centre.

With the skill, on the same layout, the agent planned to tip the three tiles, ``leaving the discard and the other tiles untouched, then take the top tile from the nearer wall stack''. In cell~13 the left arm closed on that tile at 6.5\,cm, the width the skill gives, then set it down, gripped it again at a yaw of $-90^\circ$ and stood it on end. The right arm lowered it on the row line at $x=0.320$\,m, 5\,cm beyond the outer pair tile at 0.270\,m. Opening the gripper gave success at step 375 of 600. Over the 30 test episodes, the task rises from 0 successes to 30. All 30 episodes without the skill failed the same way: the agent picked up the discard, first at steps 192--503, and never drew a tile from the wall, though two mention it in passing, once as ``the spare stack''. With the skill, all 30 drew the wall tile with the left arm and succeeded at steps 347--397.

\subsection{What the cases show}
\label{app:cases-summary}

Since the two runs of each episode differ only in the skill, the success with it shows that the agent could perceive and handle the objects of that episode. What it lacked without the skill was a fact about the task, which the skill supplied in a few lines. The facts are of the three kinds named in \cref{sec:skill-analysis}. Three are success conditions that the card leaves open: which tile to place and where it stands (make kong), which end of each pen goes up (fill pen holder) and when a moved box still counts as aligned (pack objects into box). One is a parameter that only contact reveals: a turn that pushes the nut down, with the measured heights that tell a seated nut from one on the tip (fasten screws). Two concern the ordering of steps: each pillar level is built from one pillar of each side (build tower), and the arm and pitch of the whole carry are settled by dry runs before the first slice is grasped (make toast). In five of the episodes, the agent without the skill spent its remaining steps on a repair that the skill rules out: straightening the box, turning the nut back and forth, moving the discard, taking apart the top of a tower whose lower pair could not be credited, and recovering a fallen slice while a third one took its slot. Each of these facts came from development runs in which an attempt succeeded or failed for a reason that the card does not state. The skills leave other failures in place. With the skill, fill pen holder still fails 14 times, each time with a highlighter that would not stay in the held holder; make toast fails 9 times, each time after a slice slipped; and the one failure of build tower stopped at a finished tower for a reason the record does not show. The make-toast skill also cost episodes: in three cluttered layouts, the agent with it carried each slice with one arm at a pitch of $0^\circ$ to $30^\circ$ and lost one, whereas the agent without it handed the slices between the arms, which the skill advises against, and succeeded.

\FloatBarrier
\section{Paired comparisons}
\label{sec:ablation}

\begin{table}[!htbp]
\centering
\caption{\textbf{Paired comparisons.} Each row runs the same episodes twice and changes one thing. Before and after: successes. Lost and gained: episodes that succeed only before or only after. $p$: exact two-sided sign test on these episodes. In-sample: the change was written from the records of the same episodes, so its gain is optimistic. $^a$Insert key is left out, since its skill differed between the two runs; at xhigh, sweep blocks and fill egg holder used later versions of their skills. $^b$The run without skills also kept up to 50 pictures in the conversation instead of 20. $^c$Insert key, make kong and store laptop and headphones, which passed none of their 30 episodes in an earlier run without skills, were not run again before the change and count as 0; after it they pass none either.}
\label{tab:ablation}
\small
\setlength{\tabcolsep}{3pt}
\begin{tabular}{@{}llrrrrrr@{}}
\toprule
Change & Benchmark, setting & Episodes & Before & After & Lost & Gained & $p$ \\
\midrule
\multicolumn{8}{@{}l}{\emph{Skills}} \\
Skills on & RoboDojo, 34 tasks & 1,020 & 667 & 864 & 37 & 234 & $<$0.001 \\
Skills on & RoboCasa365, 50 tasks$^b$ & 1,500 & 890 & 1,284 & 72 & 466 & $<$0.001 \\
Skills on & LIBERO-PRO, 54 tasks, validation states & 540 & 252 & 531 & 7 & 286 & $<$0.001 \\
Skills v1 $\to$ v2 (in-sample) & RoboTwin~2.0, 15 tasks, validation seeds & 150 & 126 & 142 & 7 & 23 & 0.005 \\
\midrule
\multicolumn{8}{@{}l}{\emph{Reasoning effort}} \\
High $\to$ xhigh & RoboDojo, no skills, 34 tasks$^c$ & 1,020 & 615 & 667 & 68 & 120 & $<$0.001 \\
High $\to$ xhigh & RoboDojo, skills, 33 tasks$^a$ & 990 & 782 & 844 & 54 & 116 & $<$0.001 \\
High $\to$ xhigh & RoboCasa365, skills, 50 tasks & 1,500 & 1,229 & 1,284 & 81 & 136 & $<$0.001 \\
\midrule
\multicolumn{8}{@{}l}{\emph{Harness}} \\
Picture window 20/14 $\to$ 90/60 & RoboDojo, no skills, 38 tasks & 1,140 & 805 & 818 & 96 & 109 & 0.40 \\
Remembered findings (Env\_ICL) on & RoboDojo, skills, 42 tasks & 1,260 & 973 & 981 & 91 & 99 & 0.61 \\
Re-measure after carrying & RoboDojo, no skills, 38 tasks & 1,140 & 807 & 805 & 108 & 106 & 0.95 \\
\bottomrule
\end{tabular}
\end{table}

\Cref{tab:ablation} collects the paired comparisons we ran. Each changes one thing and runs the same episodes twice.

The skills have the largest effect we measured (\cref{sec:knowledge}). At effort high on RoboDojo they gain 228 episodes and lose 60, from 615 to 783 successes. The 54 LIBERO-PRO tasks of the control are those that got a skill in the first round, 54 of the 65 tasks that failed at least one learning episode.

Raising the reasoning effort from high to xhigh helps in all three comparisons we ran. On RoboDojo without skills it gains 120 episodes and loses 68, and with skills, on 33 tasks, it gains 116 and loses 54. On RoboCasa365 with skills it gains 136 and loses 81. All three are significant ($p<0.001$). Insert key is left out of the second comparison because its skill differed between the two runs. For the same turn of the same episode it costs 2.4 to 3.3 times the reasoning tokens (\cref{sec:cost}).

Three changes to the harness, each compared at effort high, made no measurable difference. Keeping up to 90 pictures in the conversation instead of 20 gained 109 episodes and lost 96. Remembered findings (Env\_ICL), an object beside \texttt{robot} that records what earlier cells found and what went wrong, gained 99 and lost 91. Measuring an object again after it is put down, where the hand put it, gained 106 and lost 108. In each of these comparisons 15 to 19\% of the episodes changed outcome, in both directions alike. Two runs of the same episodes therefore differ in many episodes even when nothing that matters has changed, and a comparison of totals over a few hundred episodes cannot separate a small effect from this noise. Env\_ICL is off in every run of \cref{tab:main,tab:robocasa,tab:libero,tab:robotwin}.

On the code of round R0 and its 420 development episodes, we compared our system prompt with one taken from the manual of another open agent framework. Ours solved 223 episodes and the other 198, so we kept ours.

\FloatBarrier
\section{Implementation details}
\label{app:impl}

\subsection{Agent loop}
\label{app:loop}

The model has two tools: \texttt{python(code)} runs one cell, and \texttt{finish(status, summary)} ends the episode. A reply may hold several cells, which run in order. The status of \texttt{finish} gives the reason: \texttt{no\_budget} when the steps left cannot cover another attempt, \texttt{stuck} when there is no way forward, and \texttt{keep\_partial} when another attempt would put a partial result at risk. The episode ends when the benchmark ends it, when the model calls \texttt{finish}, or at a limit on replies and time: 100 replies and 90 minutes on RoboDojo, and 40 replies (100 for RoboCasa365's composite tasks) and two hours on the other benchmarks. The benchmark ends an episode on success, which it checks after every action, or when the step budget is spent. The scene is then judged as it stands. A step is one simulator action (0.04\,s of simulated time on RoboDojo).

Pictures are sent as JPEG images. The model's gateway accepts 8 to 10\,MB per request, so only a window of pictures stays in the conversation. The start pictures on the task card are always kept. When the others pass 20 pictures or 8\,MB, the oldest are replaced by a short placeholder until 14 remain. Two runs used a larger window (\cref{app:protocol}). Dropping several pictures at once keeps the cached start of the conversation valid for longer. The loop also supports Claude models and open models served with vLLM.

\subsection{Robot API conventions}
\label{app:api-conv}

Positions are in metres in the benchmark's world frame and angles in degrees. Pixels are (column, row) from the top-left corner. Pitch 0 points the hand level and 90 points it straight down. Yaw turns the hand about the vertical and roll about its own axis. \texttt{look} attaches the current pictures to the cell's result, optionally with a pixel grid, marked points, outlines, or an enlarged crop. The labels on a crop stay in full-picture pixels, so a pixel read off any picture can be passed to the other calls as it is.

A benchmark adds calls only where its robot can do more. On RoboCasa365 the base drives (\texttt{drive}), and doors and lids turn about their hinges (\texttt{arc}, \texttt{push\_arc}). A helper there picks a round dish by its rim (\texttt{rim\_pick}). LIBERO-PRO also has \texttt{arc}. Both benchmarks also hand the cell the raw depth picture (\texttt{depth}).

\subsection{Perception}
\label{app:perception}

The design grew out of the task tools of HexaShell-TaskTools (\cref{sec:v10}), which triangulated objects between calibrated RGB cameras inside each task's own code. Here one service and one set of geometry serve every task, with depth where a benchmark renders it.

One SAM~3 service \cite{carion2025sam3} runs on each machine and serves every episode there. A picture is uploaded once and then named by its hash. Words find by appearance only. \emph{Left} or \emph{second} do not pick an instance, so the model chooses among the masks by their centroids or points at the one it means.

RoboDojo has no depth camera, so \texttt{measure} works from outlines alone. A mask in one picture fixes a cone: the rays through its pixels, down to the table. To find the same object in another camera, the service is asked again, with the same words, or with clicks along the middle ray of the first outline at several heights above the table. Each candidate is scored by how well the two cones contain each other's outline, and colour breaks near ties. The object then lies in the space that every camera able to see a point shows inside its outline, a visual hull that is probed along the pixels' rays. Its top is the lowest height that still explains almost all outline pixels. Its footprint comes from the surface points each camera sees below the top. A rotated rectangle fitted to the footprint (\texttt{cv2.minAreaRect}) gives the object's size and \texttt{grip\_yaw}, the yaw that closes down-pointing fingers across its narrow side. Pixels that show the robot's own arm do not count against an outline. They are found by drawing the arm's meshes at the current joint readings. With one camera the height cannot be measured. The object is then taken to rest on its support, and the result says so and gives a wrist pose that would add a view.

\texttt{to\_world(cam, (x, y))} turns a pixel into a world point. A pixel fixes a ray, and the depth along it comes from one of three sources. The first is a horizontal plane at a height the model gives (a rim, a slot, the table). The second is the same point picked in a second picture: the two rays are intersected, and a gap of more than about 5\,mm between them reveals a mismatch \cite{hartley2004multiple}. The third is the surface under the pixel. This is the table, a measured object, or a level patch that is matched between cameras by zero-mean normalized cross-correlation over candidate heights, a plane sweep \cite{collins1996space}. Every result reports how far one pixel of picking error moves the point, and it warns when the pixel shows the robot's own arm.

LIBERO-PRO and RoboCasa365 render a depth picture with every camera. There the same calls back-project a mask's pixels through the depth picture and merge the points of all cameras that see the object. An oriented box on the floor plan gives the centre, size and grasp yaw, and the lowest and highest points give the height. Only the visible surface is measured, so when the cameras see an object from one side only, its centre leans toward them by up to half its depth. For pinch grasps, a circle fitted to the depth points around a round dish gives the outer edge of its rim. RoboTwin~2.0 renders depth as well, but in all our RoboTwin runs perception used only the colour pictures and the camera models, as on RoboDojo. On these three benchmarks the robot's own pixels come from the renderer's segmentation of the robot's parts, not from its meshes.

Measured objects keep their id for the whole episode. Measuring an object again looks for it where it was last seen, or where the gripper has taken it, and falls back on its earlier height when only one camera sees it now. The RoboDojo prompt gives the model the errors to expect. From the start pictures, low box-like objects (blocks, tiles, small flat things) usually come out within a few millimetres. Round, hollow or tall objects (balls, bowls, cups, bottles) are often 1--3\,cm off, and too high or too large rather than too low. A wrist camera held about 25\,cm above the object, together with the head camera, brings the error to 1--2\,mm. Pixels read from an enlarged crop are about twice as precise as pixels read from the whole picture.

\subsection{Motion}
\label{app:motion}

On RoboDojo, HexaShell runs as a policy inside the benchmark's own evaluation client, which asks for one action at a time. While the model thinks, the policy answers ``wait'', so the simulation stands still and no step is used. On the other three benchmarks the simulator runs in a server process with its own Python environment and talks to HexaShell over a socket.

On RoboDojo and RoboTwin~2.0 the arms are ARX X5 arms and take joint targets. HexaShell solves inverse kinematics itself, with Levenberg--Marquardt steps on the 6-D pose error of the grasp point from several starting configurations. Of the solutions that keep the arm out of the table, it takes the one with the least joint travel. A straight path is solved waypoint by waypoint, and a swing is checked for table clearance along the way. Motions follow a trapezoid-like speed profile, so long moves across the table take fewer steps while the last centimetres of a descent stay slow. A short description of the rig (where the arms stand, where the table is, how the gripper reading relates to the fingers) lets the same driver serve both benchmarks.

On LIBERO-PRO and RoboCasa365 a Franka Panda runs under robosuite's operational-space controller in delta mode. That controller lags a moving goal by about 0.16\,s and does not compensate a held load. The adapter therefore plans a straight line on one speed profile and aims each step past the plan, by feeding forward the plan's velocity and acceleration and slowly integrating what is left at rest. The same planner serves both benchmarks. On RoboCasa365 the arm also rides on a wheeled base.

\subsection{Sandbox}
\label{app:sandbox}

The model's cells run in a separate worker process with one persistent Python namespace, so variables and functions survive between cells. Every \texttt{robot} call is forwarded to the host process, which owns the arms, cameras and perception, and waits for the answer. After the worker has imported what it needs, it confines itself in three layers. Landlock limits file access to a few read-only trees (the Python installation, the HexaShell package and system libraries) and the worker's own workspace, and it blocks every TCP connection. Seccomp forbids network sockets where Landlock cannot block them, new programs and processes, and signals to or tracing of other processes. At the Python level, imports are limited to a list of modules (numpy, math, cv2, json and similar) and \texttt{open} to the workspace. These last limits give clear error messages but are not the boundary. A cell may use a limited amount of CPU time; waiting for robot calls does not count. The simulator, the task code, the layouts, the skills' sources and the network are therefore out of reach whatever a cell does.

\subsection{Writing the skills}
\label{app:skill-details}

Skills are written from the records of earlier runs and, where HexaShell-TaskTools had tools for a task, from what those tools had shown. Each claim carries its evidence. On the three newer benchmarks this is the number of episodes it was seen in, such as ``(7 eps)'', or a mark for a probe, a scripted test, a HexaShell-TaskTools tool, or ``untested''. RoboDojo skills name the development round and layouts instead, such as ``R0 8/8'' or ``(l0)''. A checking script enforces the rules of \cref{sec:skills} together with the format. On the three newer benchmarks a skill has at most 45 lines of 100 characters. On RoboTwin~2.0 the skills use only the instruction wording of the learning runs, never the unseen wording of the test. The RoboDojo skills went through four versions during the development rounds on layouts 0--9 of seed~0. The skills cite only learning and validation episodes.

\begin{table}[!htbp]
\centering
\caption{\textbf{Skill sizes.} Number of skills on each benchmark and their sizes in lines and characters.}
\label{tab:skills}
\small
\begin{tabular}{@{}lrrrrrr@{}}
\toprule
Benchmark & Tasks & Skills & Median lines & Max. lines & Median chars. & Total chars. \\
\midrule
RoboDojo & 42 & 34 & 61 & 689 & 4,270 & 178,960 \\
RoboCasa365 & 50 & 50 & 45 & 45 & 3,380 & 169,648 \\
LIBERO-PRO & 156 & 71 & 39 & 45 & 2,356 & 172,503 \\
RoboTwin~2.0 & 50 & 50 & 44 & 45 & 2,824 & 122,098 \\
\bottomrule
\end{tabular}
\end{table}

\Cref{tab:skills} gives the sizes of the skills, one per task; RoboDojo's eight Open tasks and 85 of the LIBERO-PRO tasks have none. The median skill has 39 to 61 lines and 2.4 to 4.3 thousand characters, about one page. Only three of the 205 skills contain code: a five-line loop in RoboDojo's press-by-number skill, and reference code in its insert-key and fill-egg-holder skills. The insert-key skill, with 689 lines, is the one long skill; the next longest has 138 lines.

\begin{table}[!htbp]
\centering
\caption{\textbf{Skill contents.} Share of all skill characters in each section (\%) and the evidence marks written in the skills.}
\label{tab:skill-contents}
\small
\setlength{\tabcolsep}{5pt}
\begin{tabular}{@{}lrrrr@{}}
\toprule
 & RoboDojo & RoboCasa365 & LIBERO-PRO & RoboTwin~2.0 \\
\midrule
Skills & 34 & 50 & 71 & 50 \\
Characters & 178,960 & 169,648 & 172,503 & 122,098 \\
\midrule
Flow & 31.5 & 56.0 & 33.4 & 34.2 \\
Failures / fixes & 14.7 & 17.1 & 22.5 & 21.6 \\
Judging & 9.2 & 10.5 & 21.2 & 19.8 \\
Verified parameters & 12.8 & 9.5 & 12.7 & 15.2 \\
Step costs & 5.3 & 6.3 & 9.2 & 8.4 \\
Final checks & 4.4 & -- & -- & -- \\
Clutter & 1.8 & -- & -- & -- \\
Code & 19.0 & -- & -- & -- \\
Title and other & 1.3 & 0.6 & 1.0 & 0.8 \\
\midrule
Citations (episodes) & -- & 1,136 (6,708) & 975 (7,776) & 932 (5,928) \\
Probes & -- & 6 & 513 & 0 \\
Scripted tests & -- & 50 & 0 & 0 \\
Prior-tool results & -- & 140 & 0 & 0 \\
Untested & 71 & 54 & 16 & 10 \\
\bottomrule
\end{tabular}
\end{table}

\Cref{tab:skill-contents} divides the characters of each benchmark's skills among their sections and counts the evidence marks as they are written in the skills. The largest part of every benchmark's skills is the flow, 31 to 56\% of the characters. Common failures with their fixes come next (15--23\%), then what counts as done (9--21\%), verified numbers (10--15\%) and step costs (5--9\%). The skills of the three newer benchmarks cite their evidence: 932 to 1,136 episode counts per benchmark, such as ``(7 eps)'', which name 5,928 to 7,776 episodes in all; the episodes are summed over all marks, so one episode can count several times. LIBERO-PRO's skills also cite 513 probes, short tests run for a skill, and RoboCasa365's 6 probes, 50 scripted tests and 140 results of HexaShell-TaskTools's tools. The RoboDojo skills cite development rounds and layouts instead, such as ``R0 8/8'' or ``(l0)''. Claims without evidence are marked as untested, 10 to 71 per benchmark. On RoboDojo 19\% of the characters are code, nearly all of it the reference code of insert key and fill egg holder.

\FloatBarrier
\section{Robot API reference}
\label{app:api}

\Cref{tab:api-calls,tab:api-results} give the RoboDojo API as part~3 of its prompt describes it to the model. The other benchmarks share these calls and add those named in \cref{app:api-conv}. Cameras are \texttt{head}, \texttt{left\_wrist} and \texttt{right\_wrist}, and arms are \texttt{left} and \texttt{right}. Motion calls use steps and return when the motion is done; the other calls use none. A Thing is a measured object, a Mask an outline in one picture and a Spot a world point. Every result prints as one line or a few, with a warning first when there is one (\cref{lst:results}), and its fields can be used in code. Errors are Python exceptions whose message says what to change.

\begin{table}[!htbp]
\centering
\caption{\textbf{RoboDojo API reference.} Motion consumes steps; perception and state queries do not.}
\label{tab:api-calls}
\small
\begin{tabularx}{\linewidth}{@{}>{\raggedright\arraybackslash}p{0.40\linewidth}Y>{\raggedright\arraybackslash}p{0.09\linewidth}@{}}
\toprule
Call & What it does & Returns \\
\midrule
\multicolumn{3}{@{}l}{\emph{Motion}} \\
\texttt{goto(arm, xyz, pitch, yaw, roll, grip, path=\textquotesingle auto\textquotesingle)} & Move to a pose. Path: \texttt{line}, \texttt{swing}, or \texttt{auto} (line, then a nearby swing). Optional grip starts with motion. & Move \\
\texttt{goto(arm, by=(dx, dy, dz))} & moves relative to where the hand is, or to \texttt{xyz} if given & Move \\
\texttt{goto(..., dry\_run=True)} & Preview reachability, path, step cost and proximity to arms or measured objects; no motion. & Move \\
\texttt{goto(..., contact=True)} & Contact motion; disables early stopping and \texttt{Blocked}/\texttt{Dropped} exceptions. & Move \\
\texttt{goto(..., gentle=True)} & no faster middle on long moves, for loads that slip or swing & Move \\
\texttt{gripper(arm, opening)} & open, close, or 0 to 1 (1 is 8.9\,cm between the pads) & Grip \\
\texttt{home(arm=\textquotesingle both\textquotesingle)} & lifts the hand if it is low and swings back to the start pose & Homing \\
\texttt{wait(n)} & holds everything still for $n \le 200$ steps & Waited \\
\midrule
\multicolumn{3}{@{}l}{\emph{Seeing and measuring}} \\
\texttt{look(cams, grid, marks, masks, crop, zoom, tile)} & Show images; optional grid, points, masks or enlarged crops. & Picture \\
\texttt{find(cam, words)} & outlines every instance the words describe, best first & Masks \\
\texttt{find(cam, at=(x, y))} & outlines the object under a pixel & Masks \\
\texttt{find(cam, points, labels)}, \texttt{find(cam, box)} & outline from pixels on (1) and off (0) the object, or from a box & Masks \\
\texttt{measure(target)} & Measure pose, top, size and grasp yaw. Target: text, Mask, camera/pixel or prior Thing. & Thing \\
\texttt{to\_world(cam, (x, y), z= | pair=)} & Pixel to world point via height, a second view or the visible surface. & Spot \\
\texttt{to\_pixel(xyz, cam)} & where a world point appears in a picture & pixel \\
\midrule
\multicolumn{3}{@{}l}{\emph{Reading}} \\
\texttt{task} & instruction, steps used and left, start poses & \\
\texttt{state()} & both arms: grasp point, angles, gripper, holding, distance from the start pose; steps & \\
\texttt{image(cam)}, \texttt{camera(cam)} & the current picture as an array of $480 \times 640 \times 3$; a camera's focal length, centre and pose & \\
\texttt{things} & every object measured so far, in its latest measurement & \\
\bottomrule
\end{tabularx}
\end{table}

\begin{table}[!htbp]
\centering
\caption{\textbf{API results and exceptions.} Fields are accessible in code; exceptions report the cause and recovery advice.}
\label{tab:api-results}
\small
\begin{tabularx}{\linewidth}{@{}>{\raggedright\arraybackslash}p{0.16\linewidth}Y@{}}
\toprule
Result & Fields \\
\midrule
Move & \texttt{.ok} (reached, not blocked, nothing slipped out), \texttt{.reached}, \texttt{.blocked}, \texttt{.lag\_mm} (how far the hand ended behind its command), \texttt{.dropped}, \texttt{.at}, \texttt{.angles}, \texttt{.steps} \\
Grip & \texttt{.holding} (the fingers stopped on something), \texttt{.width\_m}, \texttt{.opening} \\
Thing & \texttt{.center}, \texttt{.top}, \texttt{.size} (length, width, height), \texttt{.grip\_yaw} (the yaw that closes down-pointing fingers across the narrow side; none for round objects), \texttt{.on} and \texttt{.on\_z} (the support and its height), \texttt{.views}, \texttt{.err\_mm}, \texttt{.masks} \\
\midrule
Error & When \\
\midrule
\texttt{ValueError} & bad or impossible arguments; nothing happened \\
\texttt{NotExecuted} & a pose that cannot be reached; nothing moved, and the message offers a reachable pose \\
\texttt{Blocked} & a motion that stalled against something or ended more than 25\,mm short, or an opening that stopped on what the hand held \\
\texttt{Dropped} & what the hand held slipped out on the way \\
\texttt{NotFound} & the object to measure was not found; the message lists the candidates \\
\texttt{RobotError} & the common base of \texttt{NotExecuted}, \texttt{Blocked} and \texttt{Dropped} \\
\bottomrule
\end{tabularx}
\end{table}

\FloatBarrier
\section{Prompt and task card}
\label{app:prompt}

The RoboDojo system prompt has six parts (\cref{tab:prompt-parts}), 35 thousand characters in all. A seventh part, which explains the remembered findings of Env\_ICL, is left out in every run of this report. \Cref{lst:card} shows the task card of one episode of press by number, without its three start pictures. The task's skill follows it in the prompt (\cref{lst:note-press}).

\begin{table}[!htbp]
\centering
\caption{\textbf{RoboDojo prompt.} Six parts and their character counts.}
\label{tab:prompt-parts}
\small
\begin{tabularx}{\linewidth}{@{}lYr@{}}
\toprule
Part & What it covers & Characters \\
\midrule
1. Job and rules & Termination, judging and sandbox rules & 2,794 \\
2. World and robot & the frame, the table, the arms, the gripper and the cameras & 3,157 \\
3. The robot API & the calls, their results and errors (\cref{app:api}) & 6,522 \\
4. Perception & RGB perception workflow, active views and uncertainty & 6,553 \\
5. Motion and manipulation & Motion, grasping, contact and dual-arm coordination & 12,047 \\
6. Verification and recovery & Verification, recovery and budget & 3,919 \\
\midrule
All & & 34,992 \\
\bottomrule
\end{tabularx}
\end{table}

\begin{lstlisting}[caption={\textbf{Task card} of the episode of \cref{app:episode} (press by number, seed 0, layout 14), as the model reads it. The three start pictures follow the line that names them.},label=lst:card,captionpos=t,frame=single,columns=fullflexible]
# Task card

Instruction (from the benchmark; it is authoritative): 'Press the two red buttons the required number of times according to the number cards, then press the blue button to confirm.'
Task: press_by_number
Step budget: 700 steps for the whole episode (used so far: 0).
Start pose of each arm (the evaluator's "origin"; robot.home() returns to it):
- left: grasp point (-0.300, -0.207, 0.922), pitch -0 yaw 0 roll -0, gripper 1.00
- right: grasp point (0.300, -0.207, 0.922), pitch -0 yaw 0 roll -0, gripper 1.00
Pictures 1-3 below: head, left_wrist, right_wrist at the start; they stay in the conversation.

## How this task is scored (RoboDojo wiki)

The episode ends by itself when the last row is met. The other rows are the partial credit the scene earns if the episode ends earlier.

### Capability dimension

Memory -- Tasks requiring state tracking, sequence recall, or delayed matching.

### Description

There are two number cards, two red buttons, and one blue confirmation button.
Read the left number card, press the left red button that many times, and then
press the blue confirmation button once. Next, read the middle number card,
press the middle red button that many times, and then press the same blue
confirmation button once more. Do not finish both red buttons before pressing
the blue button. Each press is one full down-and-up of the cap, and the two
red buttons are counted separately.

### Scoring

| Score | Condition |
| --- | --- |
| 0 | The required press counts or confirm sequence is not completed exactly. |
| 100 | Button `0` is pressed exactly the number shown by `num0`, the blue confirm button is pressed, button `1` is pressed exactly the number shown by `num1`, the blue confirm button is pressed again, and the robot returns to origin. |
\end{lstlisting}

\clearpage
\section{Example skills}
\label{app:skills}

\Cref{lst:note-plug,lst:note-press,lst:note-twin} give three skills in full, as the model reads them: the two RoboDojo skills of \cref{sec:cases,app:press} and one RoboTwin~2.0 skill in the shorter format of the three newer benchmarks. Evidence in the RoboDojo skills names development rounds and layouts, such as ``R0 8/8'' or ``(l0)''; the RoboTwin~2.0 skill counts the learning and validation episodes behind each claim, such as ``(7 eps)''. ``1.0 harness'' in the press-by-number skill refers to HexaShell-TaskTools's runs of the task (\cref{sec:v10}).

\begin{lstlisting}[caption={\textbf{RoboDojo skill for plug in charger} (61 lines, no code).},label=lst:note-plug,captionpos=t,frame=single,columns=fullflexible]
# plug_in_charger

## Judging
- The card: 100 only when the charger is upright, inserted to the required depth and both arms
  are back at the start; anything not fully plugged scores 0.
- R0: letting go 3-7 mm above the depth that worked never counted (0/8). Successes ended by
  themselves during the final `robot.home()` (l0, R0a l2). The strip is loose: it may be moved.

## Flow
1. `robot.measure('charger')`, `robot.measure('power strip')` (0 steps). 'charger' NOT FOUND or
   strip-sized (it is about 5 x 4 x 2.6 cm; l1, l2, l8): measure (cam, pixel) on it. It lies
   flat, prongs toward the robot (10/10).
2. Level yaw L = the strip's grip_yaw +-90, the one in (-90, 90] (all 10 runs). Other yaws, as a
   refusal may offer, turn the prong pair across the slots. Dry-run pitch 0, roll 0, yaw L at
   the seat (below) with BOTH arms; insert with one that gets no reach note.
3. Hover the wrist nearest the strip at pitch 60, 15-20 cm up; `robot.to_world(cam, <pixel>,
   z=<strip top>)` on the middle slot of one outlet. That arm may stay there and watch (l0).
4. Grasp: ~8 cm above the charger, pitch 90, yaw = its grip_yaw, `roll=0` (else an open hand
   may arrive half a turn round: prongs up later), `grip='open'`; measure again; line down at
   its centre to 1.7 cm above the table; close ("~4.2 cm", held 24/24); lift ~12 cm.
5. One move to 5 cm above the aim point: pitch 0, roll 0, yaw L (20-30 steps). Refused as a wide
   swing (the cell stops): level above the charger first, then carry (l0: 21 + 21).
6. Line down with `contact=True` to the seat; open; line up 10 cm; look (0 steps); `robot.home()`.

## Verified parameters
- Outlet: a row of three short slots ~0.5 cm off the strip's centre line, on the far side (6/6
  zoomed views); the outer two, ~1.2 cm apart, take the prongs. Level at roll 0 after this grasp
  turns the prongs down, their pair along the strip (every run).
- Aim: grasp point short of the middle slot, along the pointing direction, by the grasp's height
  above the charger's middle (1.5 cm above the table): 0.4 cm for l0's grasp at 1.9 cm.
- Seat (centred grasp): grasp point ~3.9 cm above the table, 1.8-2 cm above the strip's measured
  top; ask 0.4 cm lower, a held load stops short (l0: 0.800 asked, 0.804 reached). A grasp moved
  along the charger's length moves the seat by as much.

## Common failures and fixes
- Opened 3-10 mm above the seat: it fell over (R0 8/8, 3 of them level; R0a l0, l1, l9, level).
  Once 4 mm above still counted (R0a l2). Push to the seat first.
- An "N deg off" note means the charger hangs tilted: never in at 3 deg or more (0/13). The arm near
  its own base got one in 7 of 8 runs; the arm reaching across was exact at the seat in 3 of 6
  (1 had 3 deg, 2 refused). Dry-run both arms (2 runs never tried the other) and the other
  outlets; else move the strip: the other arm grips it from above at its grip_yaw, 1.8 cm above
  the table, lifts 9 cm, sets it down flat nearer the middle, opens, homes; measure it again and
  re-aim (R0a l2: ~75 steps, then in level, 1/1). A refusal's other height does not help here.
- BLOCKED at the seat height: probably in, open (untested; l4 lifted such a plug out to look and
  later lost it). BLOCKED 1.5-2 cm higher: prongs on the top: lift 1 cm, re-aim (untested).
- After a fall, look where the prongs point. Level releases fell toward the robot (4/4), prongs
  then pointing away: a normal grasp turns them up. Grasp at yaw = grip_yaw +-180 with roll=0
  (untested), or turn it over on the table first (1 run, ~55 steps). Fallen away: grasp normally.
- Roll 180 to flip the prongs: the hand's housing lands on the strip (BLOCKED 3/3 runs).
- Stood on its prongs on the table, it fell (2/2). To change arms, set it down flat where the
  other arm reaches, hand still pointing down (untested).

## Final checks
- Before opening: the last Move says pitch 0 +-1, no note, at the seat height (or BLOCKED there).
- After the lift, `robot.look('head', crop=..., zoom=...)` (zoom <= 8): upright, body on the
  strip: `robot.home()` (keep ~40 steps for open, lift, home). Lying down: regrasp, no home.

## Typical step costs (R0; budget 400)
- Wrist hover 8-20 (across the table 26-35); approach 6-11, descend 1-6, close 5, lift 8-10;
  level + carry 20-30; seat 5-7; open 5; lift 6-8; home 15-26: about 100-120 in all.
- A failed try plus a regrasp cost 99-202 (median ~140): change what failed first.
\end{lstlisting}

\begin{lstlisting}[caption={\textbf{RoboDojo skill for press by number} (42 lines). Its only code is the press loop of \cref{app:press}.},label=lst:note-press,captionpos=t,frame=single,columns=fullflexible]
# press_by_number

## Judging
- A red button's count is the digit on the card right behind it; blue has no card. Order: the red farther from blue (the first red) x its digit, blue once, the red next to blue (the second red) x its digit, blue once more.
- Never press more than the count: an extra press ended a run at once (third blue, l6). No press "to be sure", none out of turn. A missing press fails too.
- A press = one move down onto the cap, then a separate move lifting the fingers clear (the card: one full down-and-up); never leave fingers on a cap.
- End: blue cap back up, both arms home; success arrives during the final `robot.home()` (8/8).

## Flow
1. Read both digits from the start head picture (10/10 right; a zoomed look is free).
2. Measure the three buttons from the head (`robot.measure(('head', (u, v)))`); use x, y and ignore the height warning: head x, y matched wrist-checked ones within 4 mm (27/27). Skip wrist hovers (untested; every R0 run paid 17-55 steps for them).
3. Arm on the first red's side: `goto` z 0.85 above it with pitch=90, yaw=0, roll=0, grip='close' (fully closed is right here); press loop; then `robot.home(arm)`: it must end there anyway, and parked near its button it touches the other arm working next door (3 runs).
4. Other arm: same approach over blue; press once; straight across at z 0.85 to the second red (~8 steps); press loop; back over blue; press once; `robot.home()`. This split: R0 1/1, 1.0 harness 9/10. One arm on both reds: 7 of 9 R0 runs, but the blue arm must leave and come back.
5. One cell per button, tally printed, no `break` on BLOCKED (that is the normal press):
```
for i in range(n):
    d = robot.goto(arm, (x, y, 0.80), pitch=90, yaw=0, roll=0, path='line', contact=True)
    print('presses done:', i + 1, d)
    u = robot.goto(arm, (x, y, 0.85), pitch=90, yaw=0, roll=0, path='line')
    if not u.ok: break  # failed lift = arms touching or arm thrown off (6/6): look
```

## Verified parameters
- Press to grasp point z 0.80 (fingertips ~1 cm below the cap top), whatever the measured top or centre says: scripted runs of this flow succeeded 8 of 8 (121-202 steps; 6 with blue at 0.803, 2 with all at 0.80), no fling, fingers never pried open. z 0.79 sometimes went 1.9 cm down the cap, pried the closed fingers open and flung the arm: R3 6 of 10 runs (all on the first press), scripted probes 3 of 5. z 0.785 and lower: wrist twisted 20-48 deg in 25/28 stalls, slid 4 cm twice, arm flung 3 times (41 presses).
- Lift to z 0.85 each time (8/8 successes lifted to 0.84-0.87; every cycle counted). Quick lifts to ~0.83 seem to have lost presses (l2).
- A down move to 0.80 ends at 0.802-0.805 (3 steps), BLOCKED or not: a full press (all counted in the 8 scripted runs; R1: stops near 0.80 with BLOCKED, 75 of 94 down moves).
- Pass pitch, yaw and roll on every press move: an omitted angle keeps the last twist (l1, l2, l9).

## Common failures and fixes
- Doubted a BLOCKED blue press and pressed again: third blue, episode over at step 338 (l6). Count it; never redo a press.
- Pressed at 0.783 in a blind loop: arm flung, its home took 472 steps, budget gone (l2). Press at 0.80.
- `if not d.ok: break` left a cap held; the re-press muddle cost 33-45 steps (l1, l8). Always lift.
- Idle arm parked next to the working one: arms touch, BLOCKED on approach or lift (l0, l5, l2). Home it.
- Down move ends far from the button (arm flung): stop, home that arm; that press still counted (2/2).
- Presses ending at z 0.81 do not count (l6: 3 test presses; scripted: 2 blue presses ending at 0.810 failed a run); ending at 0.807 still counted (1 run). Pictures do not show cap travel: trust the tally.

## Final checks
- Before the last blue press the printed tallies equal the two digits, and blue = 1.

## Typical step costs (budget 700)
- Approach from start with grip='close' ~18 (separate grip call +7); press ~6 at 0.80 (3 down, 3 up; ~13 at 0.79); to the next button ~8; home ~16.
- This flow took 121-202 in the scripted runs (5+2 to 8+9 presses); R0 successes used 282-494.
\end{lstlisting}

\begin{lstlisting}[caption={\textbf{RoboTwin~2.0 skill for stack bowls two} (35 lines).},label=lst:note-twin,captionpos=t,frame=single,columns=fullflexible]
# stack_bowls_two

## Judging
- Done when the bowls sit nested, one centred in the other, the lower one on the table, released;
  not when one rests on the other's rim, tilts or lies beside it. Not judged: which
  bowl is below (they are alike), where, which arm, nor setting the first one down as worded.

## Flow
1. Point at each bowl in the head picture (7 eps) or find them; measure each again from a wrist
   camera (10 eps). Each is about 13 cm wide and 5 cm tall.
2. Stack one onto the other where it stands only if one arm reaches pitch 90 about 10 cm above
   the lower bowl's top, over its matching rim spot, with no tilt note (4 eps). Where pitch 90
   was refused there, a bowl lowered onto it on a slant, tilted 4 degrees, and let go: both bowls
   left the view (1 ep).
3. Else set one down in the middle, its centre 2-11 cm farther from the robot than the hands'
   start grasp points, opening 0.6-1.6 cm above its gripping height, and measure it again from a
   wrist camera before bringing the other (14 eps).
4. Pinch a rim at pitch 90 closing across the wall (yaw 0 on a side rim, yaw 90 on the near rim),
   grasp point about 5.5 cm from the centre and 1 cm below the top; lift 8-14 cm (18 eps).
5. Lower the top bowl's grasp point over the base's centre plus the same offset to 2.5-5.5 cm
   above the base's top and open; success came one step into the open (18 eps).

## Verified parameters
- Pinches closed to 1.0-1.4 cm; the note that the bowl shifted after the lift was harmless (18 eps).

## Common failures and fixes
- With a rim pinch, the note that the gripper closed fully on the way meant the bowl had fallen
  (2 eps), and the hand was still opened at the target. Once the fallen bowl, taken for the other
  one, was let go near that empty target, partly over the other bowl: both left the view (1 ep).
- With a bowl, an arm 11 cm or more past the middle was refused (3 eps) or tilted 6 degrees, and
  the bowl slipped out on the way (1 ep). Set the base down in the middle instead (14 eps).

## Typical step costs
- Limit 900 steps. Successes with a base 181-360 steps in 6-15 replies (14 eps); in place 70-96
  in 7-11 (3 eps), 204 in 9 after a fall (1 ep). The failures gave up at 193 and 326 (2 eps).
\end{lstlisting}

\clearpage
\FloatBarrier
\section{Episode analysis}
\label{app:analysis}

This section looks inside the RoboDojo runs, which have a run without skills and the most detailed records: the two runs of \cref{tab:main} and, in a few places, the runs at effort high of \cref{tab:ablation}.

\subsection{Picture code written by the agent}
\label{app:tool-use}

\texttt{measure} takes words or a pixel itself and asks SAM~3 for the outline, so the agent seldom needs \texttt{find} first. We count the episodes in which any cell reads a raw picture (\texttt{robot.image}) or calls OpenCV. Without skills these are 10.3\% (105 of the 1,019 episodes with a record), and 60 of them belong to push T and swap T, the two tasks with T-shaped blocks. Deposit coin and insert tubes add 10 each. With skills the share is 9.8\%. The insert-key skill contains picture code (30 episodes), push T's skill suggests a colour mask for the target pad (22), and deposit coin adds 15. Without insert key the share with skills is 7.1\%. In the runs at effort high the agent wrote such code less often, in about 4\% of the episodes with or without skills.

\subsection{Endings, slips and infrastructure}
\label{app:failures}

\begin{table}[!htbp]
\centering
\caption{\textbf{Episode outcomes.} Successes and failed episodes of each run, by what ended them.}
\label{tab:endings}
\small
\setlength{\tabcolsep}{5pt}
\begin{tabular}{@{}lrrrrrrr@{}}
\toprule
 & & & & \multicolumn{3}{c}{Ended by the agent} & \\
\cmidrule(lr){5-7}
Run & Episodes & Success & Env. & Budget & Partial & Stuck & Limit \\
\midrule
RoboDojo, no skills & 1,020 & 667 & 102 & 226 & 15 & 7 & 3 \\
RoboDojo, skills & 1,020 & 864 & 28 & 99 & 16 & 10 & 3 \\
RoboCasa365, no skills & 1,500 & 890 & 215 & \multicolumn{3}{c}{393} & 2 \\
RoboCasa365, skills & 1,500 & 1,284 & 57 & \multicolumn{3}{c}{159} & 0 \\
LIBERO-PRO, skills & 1,560 & 1,509 & 38 & 12 & 0 & 1 & 0 \\
RoboTwin~2.0, skills & 750 & 726 & 3 & \multicolumn{3}{c}{14} & 7 \\
\bottomrule
\end{tabular}
\end{table}

\Cref{tab:endings} sorts the episodes of each run by how they ended. Success is reported by the benchmark's checker; the agent cannot declare it. A failed episode is ended by the benchmark (Env.), nearly always because the step budget was spent, by the limit on replies or time (Limit), or by the agent with \texttt{finish}, for one of three reasons: \texttt{no\_budget} when the steps left cannot pay for another attempt (Budget), \texttt{keep\_partial} when another attempt would put a partial result at risk (Partial), and \texttt{stuck} when it sees no way forward (Stuck). For RoboCasa365 and RoboTwin~2.0 our records give only the sum of the three. The RoboDojo rows are the two runs of \cref{tab:main} on the 34 tasks outside Open, and the RoboCasa365 rows the two runs of \cref{tab:robocasa}.

The agent ends most failed episodes itself, and seldom early. On RoboDojo with skills it ended 125 of the 156 failures with \texttt{finish}, 99 of them because the steps left could not pay for another attempt, with a median of 22 steps left. The benchmark ended 28, nearly all at the step budget, and 3 reached the limit on replies or time. Without skills the agent ended 248 of the 353 failures itself, 226 of them for lack of steps (a median of 14.5 left), and the benchmark ended 102. In both runs, failed episodes used a median of 97--99\% of their step budget and successful ones about half. Most failures are partial: 239 of the 353 without skills and 124 of the 156 with them earned part of the task's score. On RoboCasa365 and RoboTwin~2.0 the agent also ended most failures itself. On LIBERO-PRO, whose step limits are short, 38 of the 51 failures reached the limit.

We count the episodes in which a motion raised \texttt{Dropped}. Without skills, 289 of the 1,019 episodes with a record had a slip; 28\% of them still succeeded, against 80\% of the episodes without one. With skills, 153 of the 1,020 had a slip (46\% against 91\%). The skills thus make slips rarer, through grasp heights and openings that held in earlier runs, and they help the agent recover from one, but a slip still costs much of the chance of success. With skills, 82 of the 156 failures had a slip, and 68 of these come from six tasks, led by store laptop and headphones (22 of its 29 failures) and hang mugs (13 of 23). In 106 failures a motion stopped against something (\texttt{Blocked}), and 37 had neither a slip nor a block. Most of these 37 missed an exact final placement: deposit coin (10), where the coin did not go into the slot, insert key (9), and fold clothes (6), where the fold stayed uneven. Without skills, 208 of the 353 failures had a slip, 224 a block and 76 neither, and all 30 failures of store laptop and headphones had a slip.

In the run without skills, 58 episodes were started again: 49 because the simulator hung at its start (32 at an error dialog, 17 silent for 30 minutes), on GPUs shared with other users' jobs, and 9 because no episode began. An episode is started again only when a run gave no verdict, and each episode counts once, with its last finished run. Two episodes that hung three times on one machine were finished on another, and no episode was given up. One success ended without a result file and is counted from the benchmark's log. The first run with skills needed 17 restarts. In the two runs, 1.9\% and 2.1\% of the model requests failed at the gateway and were sent again, which costs time but no steps.

\subsection{Press by number}
\label{app:press}

Two number cards stand behind two red buttons. The agent must press the left red button as many times as the left card shows, then the blue button once, then the middle red button as many times as its card shows and the blue button once more, and bring both arms home. Without a skill, at effort high, it succeeded in 23 of 30 episodes. All seven failures ended suddenly while the agent was pressing, between steps 54 and 353 of the 700 allowed. A press too many ends the episode at once, and in several of these records the agent had made a test press to find the depth, or repeated presses it was unsure of. The skill has 42 lines. It says that every move down to a grasp height of 0.80\,m counts as a full press, \texttt{BLOCKED} or not, and must never be repeated. It gives the depth, since 0.79\,m sometimes pried the closed fingers open and flung the arm, and the loop below, which prints the count after every press:
\begin{lstlisting}[basicstyle=\ttfamily\scriptsize,frame=single,columns=fullflexible,keepspaces=true]
for i in range(n):
    d = robot.goto(arm, (x, y, 0.80), pitch=90, yaw=0, roll=0, path='line', contact=True)
    print('presses done:', i + 1, d)
    u = robot.goto(arm, (x, y, 0.85), pitch=90, yaw=0, roll=0, path='line')
    if not u.ok: break  # failed lift = arms touching or arm thrown off (6/6): look
\end{lstlisting}
With the skill the task passed all 30 episodes at both efforts. The failures without the skill came from counting, and the skill's rule addresses them: count every press once, and never press again to be sure. At effort xhigh, however, the agent without a skill also passed 29 of 30, so there the higher effort did most of what the rule does. The skill still shortens the episodes: at xhigh a success took a median of 162 steps with it and 247 without (153 and 265 at high). \Cref{app:episode} shows one episode with the skill at effort high, with five cells, 109 steps and 86 seconds. The skill's only code is the loop above, which the agent would otherwise write itself.

\subsection{Store tools in toolbox}
\label{app:toolbox}

This Open task has no skill. The agent must put a hammer, a wrench, a pair of pliers and a tape measure into a toolbox. It passed none of the 30 test episodes and none of the 10 development episodes of round R0. In R0, 66 of the 84 grasps that closed on a tool slipped: 46 of 48 on the hammer and the wrench, 19 of 24 on the pliers and 1 of 12 on the tape measure. The attempts that slipped took half of all steps (4,419 of 8,892). The pictures show how. The agent grips a long tool near its middle, as the prompt advises; as the arm lifts, the tool turns about the axis between the fingertips until it hangs, and then slides out between the pads. Most slips (48 of 66) came within the first six steps of a move. A general change to the prompt's parts on acting and checking and to how results print did not help (0 of 30 again). In scripted tests outside the agent, with each tool's place taken from the benchmark's demonstration video, the tape measure, the wrench and the hammer could each be placed (4 of 4, 6 of 6 and 2 of 3), but the pliers never (0 of about 45), and no scripted run placed all four (0 of 12). What this task lacks is a grasp that holds the long tools with this gripper, more than knowledge that a skill could add.

\subsection{Cost}
\label{sec:cost}

\begin{table}[!htbp]
\centering
\caption{\textbf{RoboDojo episode cost.} Means per test episode outside Open, without and with skills.}
\label{tab:cost}
\small
\begin{tabular}{@{}lrr@{}}
\toprule
 & No skills & Skills \\
\midrule
Turns & 32.7 & 27.0 \\
Robot calls & 161.6 & 134.3 \\
Pictures shown & 36.1 & 28.4 \\
Steps & 576 & 469 \\
Prompt tokens & 837k & 682k \\
\quad of which cached & 80\% & 89\% \\
Output tokens & 11,375 & 7,823 \\
\quad of which reasoning & 7,324 & 4,109 \\
Wall time, mean (median) & 1,263\,s (1,020\,s) & 591\,s (407\,s) \\
\bottomrule
\end{tabular}
\end{table}

\Cref{tab:cost} gives the mean cost of a RoboDojo episode over the 1,020 test episodes of the 34 tasks outside Open, in the two runs of \cref{tab:main}; one episode without skills has no record and is left out. Prompt tokens are summed over all requests of an episode, most of them read from the gateway's cache, and output tokens include the reasoning tokens. The skills cut the turns by 17\% and the steps by 19\%, and an episode needs fewer robot calls and pictures and 44\% fewer reasoning tokens. The agent follows a flow that worked instead of searching for one. The wall times compare only roughly, since the gateway was slower during parts of the run without skills.

Per request, xhigh reasons longer. With skills, over 21,703 pairs of requests made at the same turn of the same episode at high and at xhigh, the reasoning tokens rise 2.4 times (59.5 to 143.8 on average) and all output tokens 1.6 times, while the time per request stays about the same (16.1 and 15.6\,s). A request carries 22 to 25 thousand prompt tokens, 89--90\% of them read from the gateway's cache. Without skills (26,280 pairs) the reasoning tokens rise 3.3 times and a request takes twice as long, though the gateway was slower during parts of that run. On RoboCasa365 with skills, xhigh costs 3.7 times the reasoning tokens per episode (2,711 against 740) and 6\% more time (543 against 510\,s).

\FloatBarrier
\section{Example episode}
\label{app:episode}

\Cref{lst:episode} shows an episode of press by number from the RoboDojo run with skills at effort high, seed 0 and layout 14, the task card of \cref{lst:card}. Each cell is followed by what it printed. The model wrote five cells in five turns. It measured the three buttons from the head picture, pressed the left red button twice with the left arm, sent that arm home, pressed blue once with the right arm, pressed the middle red button once and blue again, and sent both arms home. The checker reported success 12 steps into the final \texttt{home}, after 109 of the 700 steps and 86 seconds. The episode used 68 thousand prompt tokens, 57 thousand of them cached, and 768 output tokens. After each of cells 2 to 4 the model saw the head picture, which is not shown here.

\begin{lstlisting}[caption={\textbf{An episode of press by number}: the model's five cells and what each printed. Long lines are wrapped.},label=lst:episode,captionpos=t,frame=single,columns=fullflexible]
### Cell 1
b0=robot.measure(('head',(244,266))); print(b0)
b1=robot.measure(('head',(320,266))); print(b1)
b2=robot.measure(('head',(396,266))); print(b2)
--- printed:
Thing#1 'at (244, 266)'  ! 1 view (head): height guessed - for a position within a few mm, look down on it with a wrist camera: robot.goto('left', (-0.151, -0.182, 0.950), pitch=60, yaw=0), then measure this Thing again  center (-0.151, -0.168, 0.781)  top 0.797  size 0.056x0.055x0.032 round  grip_yaw any  on table  1 view (head) +-9mm
Thing#2 'at (320, 266)'  ! 1 view (head, looking straight down on it): height unknown - for a position within a few mm, look down on it with a wrist camera: robot.goto('left', (-0.000, -0.183, 0.950), pitch=60, yaw=0), then measure this Thing again  center (-0.000, -0.169, 0.779)  top 0.793  size 0.058x0.058x0.028 round  grip_yaw any  on table  1 view (head) +-6mm
Thing#3 'at (396, 266)'  ! 1 view (head): height guessed - for a position within a few mm, look down on it with a wrist camera: robot.goto('right', (0.152, -0.182, 0.950), pitch=60, yaw=0), then measure this Thing again  center (0.152, -0.168, 0.781)  top 0.797  size 0.056x0.054x0.032  grip_yaw 0 deg  on table  1 view (head) +-9mm

### Cell 2
x0,y0=b0.center[:2]; x1,y1=b1.center[:2]; xb,yb=b2.center[:2]
print(robot.goto('left',(x0,y0,0.85),pitch=90,yaw=0,roll=0,grip='close'))
for i in range(2):
    print(robot.goto('left',(x0,y0,0.80),pitch=90,yaw=0,roll=0,path='line',contact=True))
    u=robot.goto('left',(x0,y0,0.85),pitch=90,yaw=0,roll=0,path='line'); print(u)
    print('Left red full presses:',i+1)
    if not u.ok: break
robot.look('head')
--- printed:
Move left: at (-0.151, -0.168, 0.851) pitch 90 yaw 0 roll 0  +-1 mm 0 deg  11 steps (line)  [689 steps left]
Move left: at (-0.151, -0.168, 0.804) pitch 89 yaw 0 roll 0  +-4 mm 1 deg  4 steps (line)  [685 steps left]
Move left: at (-0.151, -0.168, 0.848) pitch 90 yaw 0 roll 0  +-2 mm 0 deg  3 steps (line)  [682 steps left]
Left red full presses: 1
Move left: at (-0.151, -0.168, 0.802) pitch 90 yaw 0 roll 0  +-2 mm 0 deg  3 steps (line)  [679 steps left]
Move left: at (-0.151, -0.168, 0.848) pitch 90 yaw 0 roll 0  +-2 mm 0 deg  3 steps (line)  [676 steps left]
Left red full presses: 2
[picture attached: head]

### Cell 3
print(robot.gripper('left','open'))
print(robot.home('left'))
print(robot.goto('right',(xb,yb,0.85),pitch=90,yaw=0,roll=0,grip='close'))
print(robot.goto('right',(xb,yb,0.80),pitch=90,yaw=0,roll=0,path='line',contact=True))
print(robot.goto('right',(xb,yb,0.85),pitch=90,yaw=0,roll=0,path='line'))
print('Tally: left red=2, middle red=0, blue=1')
robot.look('head')
--- printed:
Grip left: open 1.00 (gap ~8.9 cm)  8 steps  [668 steps left]
Home: left home  14 steps  [654 steps left]
Move right: at (0.152, -0.167, 0.851) pitch 90 yaw 0 roll 0  +-1 mm 0 deg  11 steps (line)  [643 steps left]
Move right: at (0.152, -0.167, 0.803) pitch 90 yaw 0 roll 0  +-3 mm 0 deg  3 steps (line)  [640 steps left]
Move right: at (0.152, -0.168, 0.848) pitch 90 yaw 0 roll 0  +-2 mm 0 deg  3 steps (line)  [637 steps left]
Tally: left red=2, middle red=0, blue=1
[picture attached: head]

### Cell 4
print(robot.goto('right',(x1,y1,0.85),pitch=90,yaw=0,roll=0))
print(robot.goto('right',(x1,y1,0.80),pitch=90,yaw=0,roll=0,path='line',contact=True))
print(robot.goto('right',(x1,y1,0.85),pitch=90,yaw=0,roll=0,path='line'))
print('Tally: left red=2, middle red=1, blue=1')
robot.look('head')
--- printed:
Move right: at (0.000, -0.169, 0.850) pitch 90 yaw 0 roll 0  +-1 mm 0 deg  7 steps (line)  [630 steps left]
Move right: at (-0.001, -0.169, 0.803) pitch 90 yaw 0 roll 0  +-3 mm 0 deg  3 steps (line)  [627 steps left]
Move right: at (-0.000, -0.169, 0.848) pitch 90 yaw 0 roll 0  +-2 mm 0 deg  3 steps (line)  [624 steps left]
Tally: left red=2, middle red=1, blue=1
[picture attached: head]

### Cell 5
print(robot.goto('right',(xb,yb,0.85),pitch=90,yaw=0,roll=0))
print(robot.goto('right',(xb,yb,0.80),pitch=90,yaw=0,roll=0,path='line',contact=True))
print(robot.goto('right',(xb,yb,0.85),pitch=90,yaw=0,roll=0,path='line'))
print('Final tally: left red=2, middle red=1, blue=2; sequence complete')
print(robot.gripper('right','open'))
print(robot.home())
--- printed:
Move right: at (0.150, -0.168, 0.850) pitch 90 yaw 0 roll 0  +-2 mm 0 deg  6 steps (line)  [618 steps left]
Move right: at (0.151, -0.167, 0.804) pitch 89 yaw 0 roll 0  +-4 mm 1 deg  4 steps (line)  [614 steps left]
Move right: at (0.152, -0.167, 0.848) pitch 90 yaw 0 roll 0  +-2 mm 0 deg  3 steps (line)  [611 steps left]
Final tally: left red=2, middle red=1, blue=2; sequence complete
Grip right: open 1.00 (gap ~8.9 cm)  8 steps  [603 steps left]
EpisodeOver: the environment reports SUCCESS: the task is complete (during robot.home(), 12 steps into it)
\end{lstlisting}

\FloatBarrier
\section{Real-robot details}
\label{app:real}

The URAI adapter connects the API to two PiPER-X arms with a head camera and two wrist cameras, all RGB-D. URAI \cite{urai2026}, a separate service, plans and executes every motion: inverse kinematics, collision checking, time scaling and execution. For each \texttt{goto} the adapter sends URAI the target, asks for a planning-only preview of the whole trajectory, and executes it if the plan is accepted. With \texttt{path="auto"} it tries a straight line first. Otherwise it accepts a joint-space path only if the path stays within 3\,cm of the line and no more than 1\,cm below its end points. A refused plan raises \texttt{NotExecuted} as in simulation, and a grasp that closes on nothing raises \texttt{Dropped}. Steps are counted as executed time in units of 0.04\,s, so a step budget buys about as much robot time as on RoboDojo.

The adapter has no segmentation service at present, so \texttt{find}, \texttt{measure} and \texttt{things} are absent. The model reads pixels off the pictures, with a grid or an enlarged crop, and \texttt{to\_world} turns each pixel into a point. It uses the measured depth first, then a plane at a height the model gives, then a second view. The model checks a point by projecting it back with \texttt{to\_pixel} and marking it in a new picture. The robot's own pixels come from URAI's collision capsules, projected into each picture and compared with the measured depth.

The real-robot prompt keeps the six parts of \cref{sec:prompt} in 11,110 characters of English. A real robot has no checker, so \texttt{finish} has one more status, \texttt{done}. The model's report is stored with the episode. The adapter has 907 lines. The loop, the sandbox and the geometry of \texttt{to\_world} above it are the code that runs the simulated benchmarks. The adapter's 29 end-to-end tests pass against URAI's kinematic simulator.

The block tasks run as a loop in which each task leaves the scene for the next, and for the pens and plush toys a person resets the scene by hand before each episode. The block tasks also have more episodes on record than \cref{tab:real} shows. Over all finished episodes, shuffling reaches 10 of 11, language sorting 10 of 12, stacking 18 of 23 and unstacking 14 of 21. The one stacking failure in the table ended when the supervising model saw a person in a wrist camera and stopped the episode before the robot had moved, so it may be a false stop.

On all nine tasks, a second model supervises each episode and gives a verdict from fresh pictures and arm feedback.\footnote{GPT-6.1-Terra writes the code, supervised by GPT-6-Astra.} The counts come from these records; we did not review every episode by hand, and prompts and supervision rules changed between episodes. The papaya take-out count keeps two failures from a bug that gave the task to the wrong arm and one from a supervisor time-out, both since fixed.

Cover blocks runs on a third PiPER-X robot. All 53 finished episodes succeed; \cref{tab:real} counts the last ten.

\begin{figure}[t]
\centering
\includegraphics[width=\linewidth]{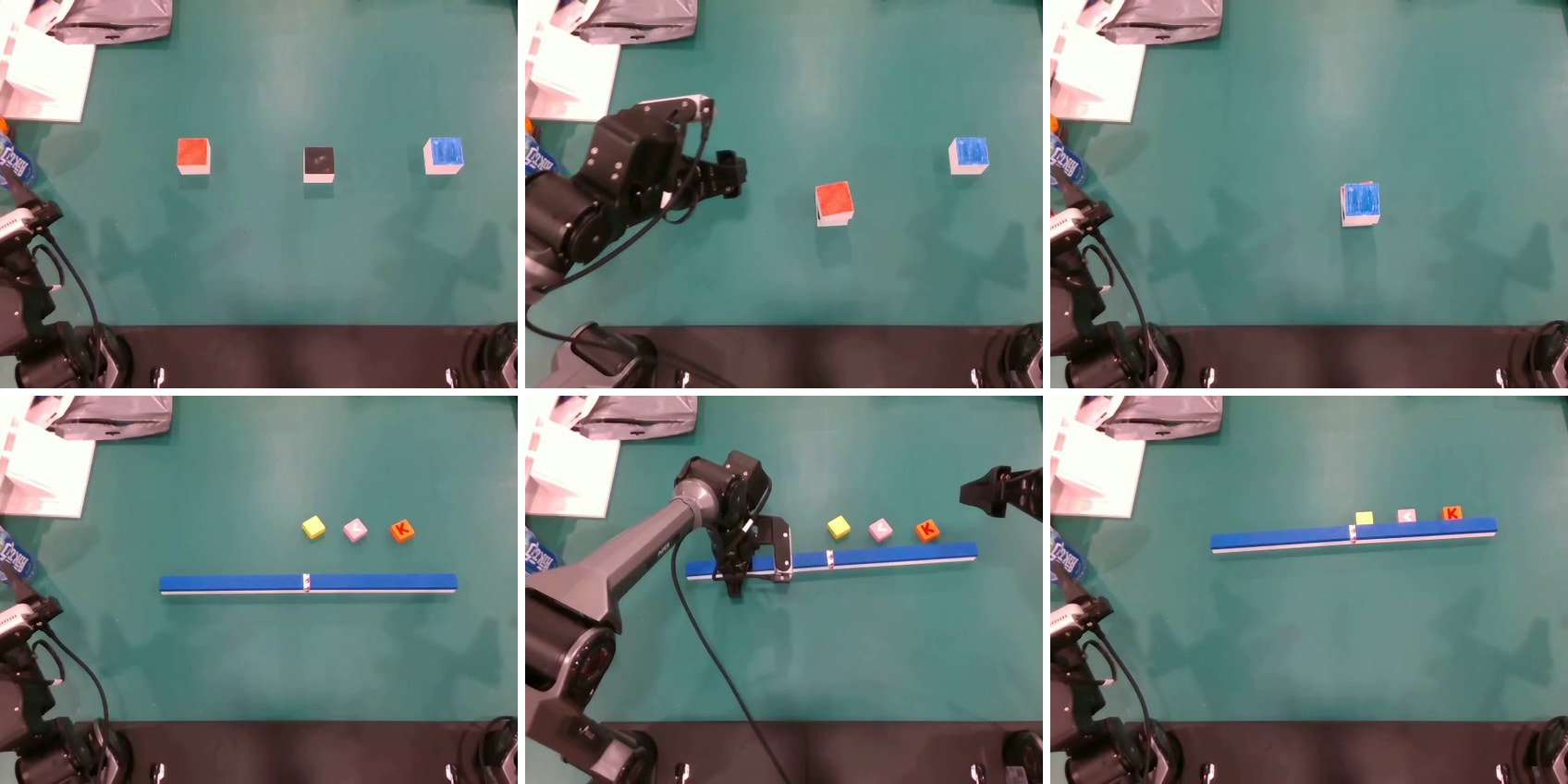}
\caption{\textbf{The PiPER-X setup}, built after RoboDojo's simulated scene and seen from above. Top: stacking three blocks into a tower. Bottom: lining up three letter blocks along a bar. From left to right: start, during the episode, end.}
\label{fig:real-setup}
\end{figure}

\Cref{fig:real-setup} shows the PiPER-X setup from above, in two episodes run earlier by HexaShell-TaskTools (\cref{sec:v10}), whose tools use only RGB pictures, the cameras' calibration and the state of the arms. \Cref{fig:real} shows one episode of HexaShell for each task.

\end{document}